\documentclass[lettersize,journal]{IEEEtran}
\pdfoutput=1
\usepackage{amsmath,amsfonts}
\usepackage{algorithmic}
\usepackage[ruled]{algorithm2e}
\usepackage{array,diagbox,subfigure}
\usepackage[caption=false,font=normalsize,labelfont=sf,textfont=sf]{subfig}
\usepackage{textcomp}
\usepackage{stfloats}
\usepackage{url}
\usepackage{verbatim,booktabs}
\usepackage{graphicx}
\usepackage{cite}
\usepackage{color,amssymb,amsthm,mathtools,enumitem,float,cleveref,multirow}
\allowdisplaybreaks[3]
\DeclareMathOperator*{\argmin}{arg\,min}

\usepackage{multirow, tabularx}
\usepackage{color,amssymb,amsmath,amsthm,mathtools,enumitem,float,subfigure}
\newcolumntype{C}{>{\centering\arraybackslash}X}

\newtheorem{lemma}{Lemma}
\newtheorem{theorem}{Theorem}
\newtheorem{definition}{Definition}

\newtheorem{proposition}{Proposition}
\newtheorem{assumption}{Assumption}

\newcommand{\y}{\mathbf{y}}
\newcommand{\x}{\mathbf{x}}
\newcommand{\W}{\mathbf{W}}
\newcommand{\Y}{\mathbf{Y}}
\newcommand{\X}{\mathbf{X}}
\newcommand{\V}{\mathbf{V}}
\newcommand{\U}{\mathbf{U}}
\newcommand{\I}{\mathbf{I}}
\newcommand{\vi}{\mathbf{v}}
\newcommand{\ui}{\mathbf{u}}
\newcommand{\Lam}{\mathbf{\Lambda}}
\newcommand{\veccommand}{\rm vec}

\usepackage{amsmath}
\usepackage{amssymb}
\usepackage{array}

\begin{document}
\title{ADMM Algorithms for Residual Network\\ Training: Convergence Analysis and\\ Parallel Implementation}
\author{Jintao~Xu$^*$,~Yifei~Li,~Wenxun~Xing
\IEEEcompsocitemizethanks{
\IEEEcompsocthanksitem Jintao Xu is with Department of Applied Mathematics, The Hong Kong Polytechnic
University, Hong Kong, China.
(jintao.xu@polyu.edu.hk)
\IEEEcompsocthanksitem Yifei Li is an independent scholar. Yifei Li is currently affiliated with Alibaba Group. This work is not associated with Alibaba Group and does not reflect the
views of the company.
(liyifei.411@outlook.com)
\IEEEcompsocthanksitem Wenxun Xing is with Department of Mathematical Sciences, Tsinghua University, Beijing 100084, China.
(wxing@mail.tsinghua.edu.cn)
\IEEEcompsocthanksitem Corresponding author: Jintao Xu
}
\thanks{This work was supported in part by the National Natural Science Foundation of China Grant No. 11771243, PolyU postdoc matching fund scheme of The Hong Kong Polytechnic University Grant No. 1-W35A, and Huawei's Collaborative Grants ``Large scale linear programming solver'' and ``Solving large scale linear programming models for production planning''.}}

\markboth{Journal of \LaTeX\ Class Files,~Vol.~xxx, No.~xxx, xxx~xxx}%
{Shell \MakeLowercase{\textit{et al.}}: A Sample Article Using IEEEtran.cls for IEEE Journals}

\maketitle

\begin{abstract}
We propose both serial and parallel proximal (linearized) alternating direction method of multipliers (ADMM) algorithms for training residual neural networks.  In contrast to backpropagation-based approaches, our methods inherently mitigate the exploding gradient issue and are well-suited for parallel and distributed training through regional updates. Theoretically, we prove that the proposed algorithms converge at an R-linear (sublinear) rate for both the iteration points and the objective function values.
These results hold without imposing stringent constraints on network width, depth, or training data size. Furthermore, we theoretically analyze our parallel/distributed ADMM algorithms, highlighting their reduced time complexity and lower per-node memory consumption. To facilitate practical deployment, we develop a control protocol for parallel ADMM implementation using Python's multiprocessing and interprocess communication. Experimental results validate the proposed ADMM algorithms, demonstrating rapid and stable convergence, improved performance, and high computational efficiency. Finally, we highlight the improved scalability and efficiency achieved by our parallel ADMM training strategy.
\end{abstract}


\section{Introduction}\label{sec:introduction}
\IEEEPARstart{T}{he} residual learning framework and its corresponding residual neural network architecture were originally proposed to facilitate effective training of deep neural networks (DNNs) for image recognition tasks \cite{He2016}. Since their inception, these techniques have rapidly become foundational paradigms within DNN design. In particular, residual connections continue to play a pivotal role in state-of-the-art deep learning applications, including large language models (LLMs). For example, residual connections are integral components of both the encoder and decoder structures in the transformer model \cite{Vaswani2017}, which underpins leading LLMs such as BERT \cite{Devlin2019} and the GPT series \cite{Brown2020,OpenAI2023}.

The expression $h(\x)+\varphi(\x)$ is termed a residual connection or residual block in neural network architectures, where $\x\in\mathbb{R}^n$, and $h, \varphi: \mathbb{R}^n\to\mathbb{R}^m$. Typically, $m=n$, and $h$ is chosen as the identity mapping. The selection of $\varphi$ determines the specific type of residual network,
for instance, conventional residual networks rely on standard convolutional operations \cite{He2016}, whereas transformer-based architectures incorporate multi-head attention mechanisms \cite{Vaswani2017}.

Neural network training remains a core research topic, attracting attention in both deep learning and optimization communities
\cite{Taylor2016,AllenZhu2019,Cui2020,JesusRubio2021,Zeng2021,Li2022,Liu2022,Wang2024}.
In this paper, we address the following \emph{multi-block nonconvex} optimization problem for training an $N$-layer residual network:
\vspace{-1.7mm}
{\small\begin{align}\label{mod:ResNet-training-model1}
\min_{\W}\frac{1}{n}\sum_{j=1}^n\ell\big(\vi_N(\x_j;\W), \y_j\big)+\lambda \Omega\big(\W\big)
\end{align}}

\vspace{-1.5mm}
\noindent
where $\W := (\veccommand(\W_1), \ldots, \veccommand(\W_N))$, the $i$th layer contains $d_i$ neurons, $\varphi_i(\x):=\sigma_i(\W_i \vi)$ with the element-wise activation function $\sigma_i: \mathbb{R} \to \mathbb{R}$ for each $i\in[N]$\footnote{For simplicity, we omit the bias term.}, the output vectors
$\vi_N\big(\x_j;\W)= f_{N}\circ \cdots\circ f_1(\x_j\big)$
with $f_i(\x) \coloneqq h_i(\x) + \varphi_i(\x)$ for $i\in[N-1]$ and $f_N(\x)\coloneqq \W_N\x$ for each $j\in [n]$, $\{(\x_j, \y_j)\}_{j=1}^n\subseteq\mathbb{R}^{d_0}\times\mathbb{R}^{d_N}$ are the given $n$ pairs of training data,
$\ell: \mathbb{R}^{d_N}\times\mathbb{R}^{d_N}\to\mathbb{R}_+$ denotes the loss function, $\Omega: \mathbb{R}^{d}\to\mathbb{R}$ denotes the regularization with $d\coloneqq  \sum_{i=0}^{N-1}d_id_{i+1}$, and parameter $\lambda>0$.

Backpropagation-based (BP-based) algorithms, including stochastic gradient descent (SGD), SGD with momentum (SGDM), and Adam \cite{Kingma2017}, are widely used to train neural networks. They calculate gradients by iteratively propagating partial derivatives throughout the network using the chain rule. However, as highlighted in \cite{Taylor2016,Choromanska2019,Gu2020}, BP-based algorithms present several notable limitations, described as follows.
\begin{enumerate}
\item When computing gradients through the chain rule, BP-based training algorithms encounter the well-known exploding gradient issue during the training of DNNs \cite{Goodfellow2016}. This phenomenon causes the magnitude of (stochastic) gradient components to progressively increase, thereby hindering effective training and potentially leading to computational failures (see \Cref{sec:experiments}).
\item To update the weight matrix in the $i$th layer, partial-derivatives with respect to all preceding layers (i.e., layers $i-1$ through $1$) must first be computed. Due to the inherently sequential nature of this gradient computation, BP-based training methods do not readily lend themselves to parallelization across layers
    \cite{CarreiraPerpinan2014}.
\item In distributed training scenarios, the scalability of BP-based methods is  inherently limited. Specifically, the sequential dependency induced by the chain rule restricts layer-wise partitioning across multiple computational nodes. Consequently, each node typically maintains a complete copy of the network weights, leading to significant memory and storage overhead.
\end{enumerate}
The alternating direction method of multipliers (ADMM) \cite{Han2022} is a widely adopted optimization technique that updates primal variables by solving subproblems derived from the augmented Lagrangian, followed by dual variable updates in each iteration. To address the limitations inherent to BP-based methods, we propose a family of serial and parallel ADMM-based training algorithms specifically designed for residual neural networks, herein referred to as \textbf{\emph{RADMM}}. Our approach leverages the structural properties of residual networks, thereby effectively circumventing the sequential dependencies imposed by the chain rule. \emph{Consequently, the proposed RADMM framework mitigates the above drawbacks of BP-based training as below.}
\begin{enumerate}
\item By eliminating gradient computations based on the chain rule, the proposed approach effectively addresses the exploding
    gradient issues commonly encountered in BP-based methods.
\item By leveraging the proposed parallel regional update (\textbf{\emph{PRU}}) strategy, our algorithms naturally facilitate parallel execution across network layers (see \Cref{subsec:parallel admms}), significantly reducing computational time.
\item
By enabling each computing node to store only the variables associated with its corresponding block, our proposed approach significantly reduces per-node memory requirements during distributed training.
\end{enumerate}
To facilitate the development of parallelizable proximal (linearized) RADMM algorithms (see \Cref{def:proximal} for the proximal (linearized) operator), we first reformulate problem (\ref{mod:ResNet-training-model1}) into an equivalent constraint multi-block nonconvex optimization formulation that fully exploits the inherent structural properties of residual networks, as follows.
\vspace{-1.5mm}
{\small\begin{align}
\small
\min_{\W}&~\mathcal{J} := \frac{1}{n}\sum_{j=1}^n\ell\big(\vi_N^j, \y_j\big)+\lambda \Omega\big(\W\big)\nonumber\\
\mbox{s.t.}~
& \vi_{i}^j = h_i\big(\vi_{i-1}^j\big) + \sigma_i\circ g_i\big(\vi_{i-1}^j; \W_i\big), i\in[N-1], j\in[n],\nonumber\\
& \vi_{N}^j=\W_{N}\vi_{N-1}^j, j\in[n],\label{mod:ResNet-training-model2}
\end{align}}

\vspace{-1mm}
\noindent where $\vi_0^j = \x_j$ for each $j\in[n]$, and
$\W_i, i=1, \ldots, N$ are tightly coupled, significantly hindering efficient and parallel training of residual networks.

By introducing a set of auxiliary decision variables $\{\vi_i^j\}$, we derive the following relaxation model termed the two-splitting relaxation, which forms the basis for designing two-splitting proximal (linearized) RADMMs (see \Cref{sec:2-splitting-ADMM}):
\vspace{-1.5mm}
{\small\begin{align}
\min_{\substack{\W, \vi}}&~
\mathcal{J}+\frac{\mu}{2}\sum_{j=1}^n\sum_{i=1}^{N-1}\Vert h_i(\vi_{i-1}^j) + \sigma_i\circ g_i\big(\vi_{i-1}^j; \W_i\big)-\vi_{i}^j\Vert^2\nonumber\\
\mbox{s.t.}~
&~\vi_{N}^j=\W_{N}\vi_{N-1}^j, j\in[n],\label{mod:ResNet-training-model4}
\end{align}}

\vspace{-1.5mm}
\noindent where $\vi := (\vi_i^j)$, and $\mu>0$. Furthermore, to improve training efficiency of deeper residual networks, as suggested in \cite{Zeng2019}, we introduce an additional set of auxiliary decision variables $\{\ui_i^j\}$.
Consequently, the network can be reformulated as:
\vspace{-1mm}
{\small\begin{align*}
\begin{aligned}
&\vi_{i}^j = h_i(\vi_{i-1}^j) + \sigma_i(\ui_i^j); \\
& \ui_i^j = g_i\big(\vi_{i-1}^j; \W_i\big), i\in[N-1], j\in[n].\nonumber
\end{aligned}
\end{align*}}
\vspace{-3mm}

\noindent
Then we derive the following relaxation, termed the three-splitting
relaxation, which serves as the foundation for developing the three-splitting proximal (linearized) RADMM algorithms (see \Cref{sec:3-splitting-ADMM}):

\vspace{-4.5mm}
{\small\begin{align}
\min_{\W, \ui, \vi}&~
\mathcal{J}+\frac{\mu}{2}\sum_{j=1}^{n}\sum_{i=1}^{N-1}\Vert h_i(\vi_{i-1}^j) + \sigma_i(\ui_i^j)-\vi_{i}^j\Vert^2\nonumber\\
\mbox{s.t.}~
&~\ui_i^j = g_i\big(\vi_{i-1}^j; \W_i\big), i\in[N-1], j\in[n],\nonumber\\
&~\vi_{N}^j=\W_{N}\vi_{N-1}^j, j\in[n],\label{mod:ResNet-training-model5}
\end{align}}

\vspace{-1mm}
\noindent where $\ui := (\ui_i^j)$.

\begin{figure}[h]
  \centering
  \includegraphics[width=1\linewidth]{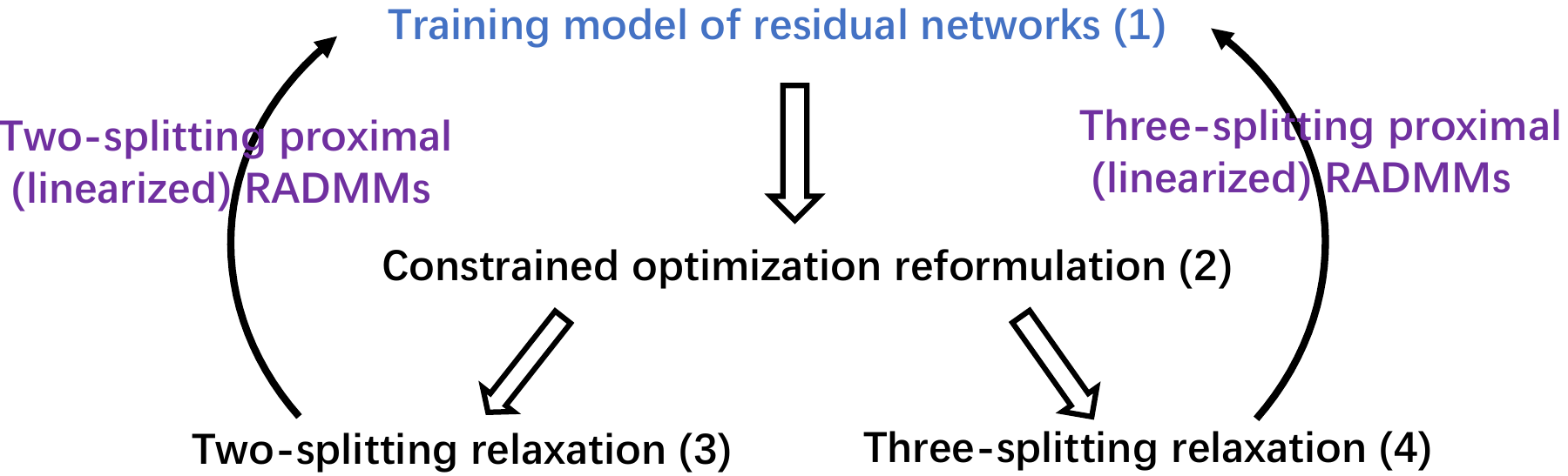}
\caption{Relationships between training problem, relaxations, and RADMMs.}
		\label{fig:relationship-relaxation-admm}
\end{figure}

For simplicity, we adopt the mean squared error (MSE) loss for $\ell$, employ the ridge regression penalty for $\Omega$, and set $h_i$ as the identity mapping. Accordingly, problem (\ref{mod:ResNet-training-model2}) can be reformulated as the following matrix optimization problem:

\vspace{-5mm}
{\small\begin{align}
\min_{\W}~& \mathcal{J}\big(\W, \V_N\big):=\frac{1}{2}\Vert \V_{N}-\Y\Vert^{2}+\frac{\lambda}{2}\sum_{i=1}^{N}\Vert \W_{i}\Vert^{2}\nonumber\\
\mbox{s.t.}~
& \V_{i}=\V_{i-1}+\sigma_i\big(\W_{i}\V_{i-1}\big), i\in[N-1],\nonumber\\
& \V_{N}=\W_{N}\V_{N-1},\label{mod:RTM1}
\end{align}}
\vspace{-6mm}

\noindent where $\X\coloneqq(\x_1, \ldots, \x_n)\in\mathbb{R}^{d_0\times n}$, $\Y\coloneqq(\y_1, \ldots, \y_n)\in\mathbb{R}^{d_N\times n}$, $\V_i\coloneqq(\vi_i^1, \ldots, \vi_i^n)\in\mathbb{R}^{d_0\times n}$, $i\in[N-1]$ and $\V_N\coloneqq(\vi_N^1, \ldots, \vi_N^n)\in\mathbb{R}^{d_N\times n}$ are the output/input matrices of each layer. It should be noted that (\ref{mod:RTM1}) is also a multi-block nonconvex optimization problem. As discussed in \Cref{sec:convergence-of-2-splitting-ADMM} and \Cref{sec:convergence-of-3-Splitting-ADMM}, the theoretical analysis of the convergence of RADMMs is challenging. \Cref{fig:relationship-relaxation-admm} illustrates the relationships among the training problem, the relaxation models, and the proximal (linearized) RADMM algorithms.
\vspace{-1mm}
\vspace{-1mm}
The main contributions of this paper are summarized as below.
\begin{enumerate}
  \item Algorithmically, we develop parallelizable proximal (linearized) ADMM training methods for residual networks (RADMMs) by exploiting their intrinsic structure, and incorporating auxiliary variables. Moreover, we implement our methods in parallel and distributed architectures via the Parallel Regional Update (PRU) approach.
\item Theoretically, we establish the convergence of both the iteration points and objective function values for the proposed serial and parallel proximal RADMM algorithms, without imposing restrictive assumptions on network width, depth, or dataset size. Furthermore, we prove that these algorithms exhibit an R-linear/sublinear convergence rate based on the value of the KL exponent.
  \item Theoretically, we demonstrate that the proposed parallel strategy significantly reduces time complexity under conditions of low communication overhead. Additionally, our distributed approach reduces the per-node runtime memory requirement from cubic to quadratic.
\item Empirically, the proposed RADMM algorithms exhibit faster and more stable convergence, improved performance, and higher training speed compared to SGD, SGDM, and Adam. Furthermore, we have developed and implemented a control protocol for parallel RADMM training using Python's multiprocessing and interprocess communication (IPC) mechanisms, demonstrating substantial advantages derived from parallelization.

\end{enumerate}
The remainder of this paper is organized as follows. We start with the related work in Section \ref{sec:relatedwork}. Notations and definitions are shown in Section \ref{sec:preliminaries}. Next, two- and three-splitting RADMMs and their convergence analysis are presented in Section \ref{sec:2-splitting-ADMM} and Section \ref{sec:3-splitting-ADMM}, respectively. Parallel, distributed algorithms and their advantages are shown in Section \ref{sec:advantages-of-parallel-implementation}. Experiments are reported in Section \ref{sec:experiments}. Finally, we make some concluding remarks in Section \ref{sec:concluding_remarks}.

\section{Related Work}\label{sec:relatedwork}
\subsection{Alternatives to BP-Based Training Algorithms}
Algorithmically, Taylor et al. \cite{Taylor2016}, Zeng et al. \cite{Zeng2021}, Wang et
al. \cite{Wang2019}, and Wang et al. \cite{Wang2020} developed ADMM training algorithms for feedforward neural networks (FNNs). Moreover,
this approach has been applied to train graph-augmented
multilayer perceptrons \cite{Wang2024}. Gu et al. \cite{Gu2020} introduced a class
of lifted relaxation models known as Fenchel lifted network.
Li et al. \cite{Li2022} proposed the lifted proximal operator machine
for FNN training. Li et al. \cite{Li2022}, Gu et al. \cite{Gu2020}, and Zeng et
al. \cite{Zeng2019} developed block coordinate descent (BCD) training
algorithms for FNNs. Cui et al. \cite{Cui2020} designed a majorization-
minimization algorithm for an exact reformulation of the
DNN training optimization problem. Liu et al. \cite{Liu2022}, \cite{Liu2023}
developed a smoothing proximal gradient algorithm and an
inexact augmented Lagrangian algorithm for autoencoders and
leaky ReLU DNNs with group sparsity training, respectively.

Theoretically, Cui et al. \cite{Cui2020} demonstrated that any accumulation point of the sequence generated by the majorization-minimization training algorithm is a directional-stationary point. For the autoencoders training problem and related models, Liu et al. \cite{Liu2022} investigated the relationships between global (or local) minimizers and (generalized) directional-stationary points. Moreover, for leaky ReLU DNNs with group sparsity training, Liu et al. \cite{Liu2023} discussed the relationship between global (or local) minimizers and certain stationary points. Convergence of the BCD training algorithms is studied by \cite{Li2022,Zeng2019}. Zeng et al. \cite{Zeng2021} established both the convergence and an $\mathcal{O}(1/k)$ convergence rate for ADMM training algorithms applied to FNNs.  Additionally, Xu et al. \cite{Xu2022} provided a convergence rate analysis framework for a general class of alternating minimization-type DNNs training methods.
\subsection{Alternating Direction Method of Multipliers}
ADMM has attracted significant attention in the machine learning community \cite{Taylor2016,Wang2019,Lai2020,Wang2020,Zeng2021,Deng2023,Wang2024}.
It is well established that (proximal) ADMM converges for two-block separable convex optimization problems subject to linear constraints \cite{Fazel2013,Han2022}. Moreover, by imposing additional assumptions, convergence results have been derived for the (semi-proximal) ADMM when applied to multi-block convex optimization problems with linear constraints
\cite{Li2016,Hong2017} as well as
multi-block nonconvex optimization problems with linear constraints
\cite{Yang2017,Wang2019a}.

\subsection{DNNs Parallel Training}
Parallel and distributed computing are widely employed in machine learning research.
In the context of DNN training, two prevalent strategies are model parallelism and data parallelism \cite{Harlap2018,Wang2020,Lee2022,Li2022,Wang2024}. With model parallelism, the neural network is partitioned into several segments, each trained on a separate processor.
In contrast, data parallelism distributes subsets of the training data across processors, with each processor independently training the entire model using its local data.

\section{Preliminaries}\label{sec:preliminaries}
\subsection{Notations}
$\Vert\cdot\Vert$ denotes the $l_2$ norm for vectors and the Frobenius norm for matrices, respectively. $\odot$ denotes the Hadamard product, while $\varliminf f$ denotes the limit inferior of $f$. $\overline{v}$ and $\underline{v}$ denote the upper and lower bounds of variable $v$, respectively.
Additionally, $\veccommand(\X)$ denotes the vectorization of matrix $\X$. $\langle \X, \Y\rangle = {\rm tr }(\X\Y^{\mathrm{T}})$ for matrices $\X$ and $\Y$.

\subsection{Optimization and Variational Analysis}
\begin{definition}[\cite{Bertsekas2015}]\label{def:proximal}
Let $f$ be a differentiable closed proper convex function. Then the mapping
\vspace{-1mm}
{\small\begin{align*}
\text{\rm prox}_{\frac{1}{\alpha}f}(\bar\x)=\argmin_{\x}\Big\{f(\x)+\frac{\alpha}{2}\Vert \x-\bar\x\Vert^{2}\Big\}
\end{align*}}
\vspace{-3mm}

\noindent with $\alpha>0$, is said to be the proximal operator.
And we also have the linearized version:
\vspace{-1mm}
{\small\begin{align*}
\argmin_{\x}\Big\{\langle \nabla f(\bar\x), \x-\bar\x\rangle+\frac{\alpha}{2}\Vert \x-\bar\x\Vert^{2}\Big\}.
\end{align*}}
\vspace{-7mm}
\end{definition}
The history of Kurdyka-\L ojasiewicz (KL) property goes back to the work of Kurdyka \cite{Kurdyka1998} and \L ojasiewicz \cite{Lojasiewicz1963,Lojasiewicz1993}. We first present the next two definitions.
\begin{definition}[\cite{Mordukhovich2006,Rockafellar1998}]\label{def:Frechet subdifferential}
The Fr{\'{e}}chet subdifferential $\widehat{\partial}f(\x)$ of $f$ at $\x\in {\rm dom} f$ is the following set
\vspace{-1.5mm}
{\small\begin{equation*}\small
\Bigg\{\vi\left|\varliminf_{\y\to \x}\frac{f(\y)-f(\x)-\langle \vi, \y-\x\rangle}{\Vert \y -\x\Vert}\geq0\right.\Bigg\}.
\end{equation*}}
\end{definition}
\begin{definition}[\cite{Mordukhovich2006,Rockafellar1998}]\label{def:limiting subdifferential}
The limiting subdifferential $\partial f(\x)$ of $f$ at $\x\in {\rm dom} f$ is the following set
\vspace{-1.5mm}
{\small\begin{equation*}
\big\{\vi\big|\exists~\x^{k}\to \x, f(\x^{k})\to f(\x), \vi^{k}\to \vi, \vi^{k}\in\widehat{\partial}f(\x^{k})\big\}.
\end{equation*}}
\vspace{-7mm}
\end{definition}
\vspace{-2mm}
\begin{definition}[\cite{Bertsekas2015}]
A function $f$ is called as proper if $f(\x)>-\infty$ for every $\x$ and $f(\x)<\infty$ for at least one $\x$, and lower semicontinuous at $\x$ if $f(\x) \le \varliminf_{k\to\infty} f(\x_k)$ for every sequence $\{\x_k\}$ satisfying $\x_k\to\x$.
\end{definition}
\vspace{-3mm}
\begin{definition}[\cite{Attouch2010,Li2018}]
A proper lower semicontinuous function $f$ is said to have the Kurdyka-\L ojasiewicz (KL) property at $\x^*$ with exponent $\theta$ if there exist $c\in(0, +\infty)$, $\tau\in(0, +\infty]$, $\theta\in[0, 1)$ and an open set $\mathcal{N}_{\x^*}$ containing $\x^*$ such that $(f(\x)-f(\x^*))^{\theta}\leq c\inf_{\vi\in\partial f(\x)}\Vert \vi\Vert$
for all $\x\in \mathcal{N}_{\x^*}\cap\{\x|f(\x^*)<f(\x)<f(\x^*)+\tau\}$.
Parameter $\theta$ is said to be KL exponent at $\x^*$.
\end{definition}
\begin{definition}[\cite{Krantz2002}]
A function $f$, defined on an open set $\mathcal{U}\subseteq\mathbb{R}$, is called as real analytic on a subset $\mathcal{S}\subseteq\mathcal{U}$ if for every point $x\in\mathcal{S}$, there exists an interval $\mathcal{N}_x\subseteq\mathcal{U}$ centered at $x$ such that
$f(z) = \sum_{i=0}^\infty a_i(z-x)^i$ for all $z\in \mathcal{N}_x$.
\end{definition}
\vspace{-1.5mm}
In this paper, we assume that the activation $\sigma_i$ for $i\in[N-1]$ are all real analytic such that there exist $\psi_0$, $\psi_1$, and $\psi_2$ satisfying $\vert\sigma_i(x)\vert\leq \psi_{0}$, $\vert \sigma_i^{'}(x)\vert\leq\psi_{1}$, and $\vert\sigma_i^{''}(x)\vert\leq \psi_{2}$.
These assumptions are satisfied by the sigmoid, hyperbolic tangent, sine, and cosine functions.
Notably, we do not impose any restrictions on the network depth $N\in\mathbb{N}_+$, the width $d_i\in\mathbb{N}_+$ of each layer, or the number $n$ of samples.

\section{Two-splitting RADMMs}\label{sec:2-splitting-ADMM}
In this section, we introduce scalable two-splitting proximal (linearized) RADMM algorithms and provide their rigorous theoretical analysis.

\subsection{Two-Splitting Proximal (Linearized) RADMMs}
Our two-splitting RADMMs are designed based on the following constraint reformulation of (\ref{mod:ResNet-training-model4}) with MSE loss, ridge regression penalty, and identity mapping $h_i$, $i\in[N-1]$:
\vspace{-2mm}
{\footnotesize\begin{align}
~\min_{\W, \V}~& \mathcal{J}\big(\W, \V_N\big)+\frac{\mu}{2}\sum_{i=1}^{N-1}\Vert \V_{i-1}+\sigma_i\big(\W_{i}\V_{i-1}\big)-\V_{i}\Vert^{2}\nonumber\\
\mbox{s.t.}~
& \V_{N}=\W_{N}\V_{N-1},\label{mod:2SARTM}
\end{align}}
\vspace{-5mm}

\noindent where $\V_i=(\vi_i^1, \ldots, \vi_i^n)$ for each $i\in [N]$, $\V := (\veccommand(\V_1),$\\$ \ldots, \veccommand(\V_N))$, and its augmented Lagrangian function:
\vspace{-1.5mm}
{\footnotesize\begin{align}
\mathcal{L}_{\beta}^{2s}\big(\Psi_{2s}\big)
&\coloneqq\mathcal{J}\big(\W, \V_N\big)+\frac{\mu}{2}\sum_{i=1}^{N-1}\Vert \V_{i-1}+\sigma_i\big(\W_{i}\V_{i-1}\big)-\V_{i}\Vert^{2}\nonumber\\
&+\langle\Lam, \W_{N}\V_{N-1}-\V_{N}\rangle+\frac{\beta}{2}\Vert \W_{N}\V_{N-1}-\V_{N}\Vert^{2},\label{eq:2SARTM-ALF}
\end{align}}
\vspace{-4mm}

\noindent in which $\Psi_{2s} := (\{\veccommand(\W_i)\}_{i=1}^N, \{\veccommand(\V_i)\}_{i=1}^N, \veccommand(\Lam))$, $\Lam\in\mathbb{R}^{d_N\times n}$ is the dual variable, parameter $\beta>0$.
Update subproblems of each block variable based on (\ref{eq:2SARTM-ALF}) in a Gauss-Seidel manner are listed in \Cref{table:updates 2-spitting ADMM}, in which ``P'' (``PL'') represents the proximal (linearized) update, espectively. Based on \Cref{table:updates 2-spitting ADMM}, the pseudocode of two-splitting RADMM training algorithms are given in Algorithm \ref{alg:2sADMM_s}.
{\footnotesize\begin{table}[h]
\caption{Updates of block variables in two-splitting RADMMs.}
\renewcommand\arraystretch{0.01}
\centering\label{table:updates 2-spitting ADMM}
    \begin{tabularx}{0.489\textwidth}{p{0.34in}| p{2.8in}}
        \hline
 Variables &~~~~~~~~~~~~~~~~~~~~~~~~~~~~~~~Updates\\
        \hline
\begin{align*}\W_N^{k+1}\end{align*} & \begin{align}\label{eq:2s-Wn-update}
(\beta \V_{N}^{k}-\Lam^{k})\V_{N-1}^{k~\mathrm{T}}(\lambda \I+\beta \V_{N-1}^{k}{\V_{N-1}^{k~\mathrm{T}}})^{-1}~~~~~~~~
\end{align}\\
\hline
    \multirow{1200}{=}{\begin{equation*}\W_i^{k+1}\end{equation*}}
 &   \begin{align}\label{eq:2s-prox-Wi-update}
                        \text{P:}~~\argmin\limits_{\Vert \W_i\Vert\leq\varkappa}\big\{\frac{\lambda}{2}\Vert \W_{i}\Vert^{2}+\frac{\omega_{i}^{k}}{2}\Vert \W_{i}-\W_{i}^{k}\Vert^{2}~~~~~~~~~\nonumber\\
+\frac{\mu}{2}\Vert \V_{i-1}^{k}+\sigma_i(\W_{i}\V_{i-1}^{k})-\V_{i}^{k}\Vert^{2}\big\}~~~~~~~~~~~
                        \end{align} \\
        \cline{2-2}
  & \begin{align}
                        \label{eq:2s-proxgra-Wi-update}
\text{PL:}~~-\frac{\mu}{\lambda+\tau_{i}^{k}}[(\V_{i-1}^{k}
+\sigma_i(\W_{i}^{k}\V_{i-1}^{k})-\V_{i}^{k})\odot\nonumber~~\\
\sigma_i^{'}(\W_{i}^{k}\V_{i-1}^{k})]\V_{i-1}^{k~\mathrm{T}}+\frac{\tau_{i}^{k}}{\lambda+\tau_{i}^{k}}\W_{i}^{k}~~~~~~~~~~~ \end{align} \\
        \hline
            \multirow{2000}{=}{\begin{equation*}\V_i^{k+1}\end{equation*}}
 &   \begin{align}\label{eq:2s-prox-Vi-update}
                        \text{P:}~~\argmin\limits_{\V_{i}}\{\frac{\mu}{2}\Vert \V_{i}+\sigma_{i+1}(\W_{i+1}^{k+1}\V_{i})-\V_{i+1}^{k}\Vert_{F}^{2}\nonumber\\
 \frac{\mu}{2}\Vert \V_{i-1}^{k+1}+\sigma_i(\W_{i}^{k+1}\V_{i-1}^{k+1})-\V_{i}\Vert_{F}^{2}~~~~~~~~~~\nonumber\\
                        +\frac{\nu_{i}^{k}}{2}\Vert \V_{i}-\V_{i}^{k}\Vert_{F}^{2}\}~~~~~~~~~~~~~~~~~~~~~~~~~~~~~~~~~
 \end{align} \\
        \cline{2-2}
  & \begin{align}
     \label{eq:2s-proxgra-Vi-update}
\text{PL:}~~\frac{\mu}{\mu+\iota_{i}^{k}}[\sigma_i(\W_{i}^{k+1}\V_{i-1}^{k+1})+\V_{i-1}^{k+1}~~~~~~~~~~~~~~\nonumber\\
+\V_{i+1}^{k}
-\V_{i}^{k}-\sigma_{i+1}(\W_{i+1}^{k+1}\V_{i}^{k})]~~~~~~~~~~~~~\nonumber\\
+\frac{\iota_{i}^{k}}{\mu+\iota_{i}^{k}}\V_{i}^{k}
-\frac{\mu}{\mu+\iota_{i}^{k}}{\W_{i+1}^{k+1~\mathrm{T}}}
[(\V_{i}^{k}-\V_{i+1}^{k}\nonumber\\
+\sigma_{i+1}(\W_{i+1}^{k+1}\V_{i}^{k}))\odot\sigma_{i+1}^{'}(\W_{i+1}^{k+1}\V_{i}^{k})]~~~~~~
                    \end{align} \\
\hline
\begin{align*}\\ \\\V_{N-1}^{k+1}\end{align*}&
\begin{align}\label{eq:2s-Vn-1-update}
\mu(\mu \I+\beta\W_{N}^{k+1~\mathrm{T}}\W_{N}^{k+1})^{-1}~~~~~~~~~~~~~~~~~~~~~~\nonumber\\
(\sigma_{N-1}(\W_{N-1}^{k+1}\V_{N-2}^{k+1})+\V_{N-2}^{k+1})+~~~~~~~~~~~~~~\nonumber\\
\beta(\mu \I+\beta\W_{N}^{k+1~\mathrm{T}}\W_{N}^{k+1})^{-1}\W_{N}^{k+1~\mathrm{T}}\V_{N}^{k}~~~~~\nonumber\\
-(\mu \I+\beta\W_{N}^{k+1~\mathrm{T}}\W_{N}^{k+1})^{-1}\W_{N}^{k+1~\mathrm{T}}\Lam^{k}~~~~~~
\end{align}\\
\hline
\begin{align*}\V_{N}^{k+1}\end{align*} &\begin{align}\label{eq:2s-Vn-update}\frac{1}{1+\beta}\big(\Y+\beta \W_{N}^{k+1}\V_{N-1}^{k+1}+\Lam^{k}\big)~~~~~~~~~~~~~~~\end{align}\\
\hline
\begin{align*}\Lam^{k+1}\end{align*} &
\begin{align}\label{eq:2s-lambda-update}\Lam^{k}+\beta\big(\W_{N}^{k+1}\V_{N-1}^{k+1}-\V_{N}^{k+1}\big)~~~~~~~~~~~~~~~~~\end{align}
\\
  \hline
    \end{tabularx}
\end{table}}
{\footnotesize\begin{algorithm}[h]\label{alg:2sADMM_s}
\footnotesize
\SetKwInput{Initialization}{Initialization}
\SetKw{Return}{return}
\SetKwFor{ForParallel}{parallel\_for}{do}{end}
\SetKwFunction{Range}{range}
\SetKwInput{Input}{Input}
\SetKwInput{Output}{Output}
\caption{Two-Splitting RADMMs}
\Input{$\X, \Y, K, \lambda, \mu, \beta, \{\omega_i^k\}_{i=1}^{N-1}$, $\{\nu_i^k\}_{i=1}^{N-1}$, and $\{\W_{i}^{0}\}_{i=1}^N$.}
$\V_{0}^{k}\gets \X$, $\V_{i}^{0}\gets \V_{i-1}^{0}+\sigma_i(\W_{i}^{0}\V_{i-1}^{0})$, $i\in[N-1]$, $\V_{N}^{0}\gets \W_{N}^{0}\V_{N-1}^{0}$, $\Lam^{0}\gets \boldsymbol{0}$.\\
\For{$k\gets0$~{\rm\textbf{to}}~$K-1$}{
Update $\W_{N}^{k+1}$ by iteration equation (\ref{eq:2s-Wn-update}).\\
\For{$i\gets N-1$~{\rm\textbf{to}}~$1$}{
Update $\W_{i}^{k+1}$ by solving (\ref{eq:2s-prox-Wi-update}) in proximal RADMM (by iteration equation (\ref{eq:2s-proxgra-Wi-update}) in proximal linearized RADMM).}
\For{$i\gets1$~{\rm\textbf{to}}~$N-2$}{
Update $\V_{i}^{k+1}$ by solving (\ref{eq:2s-prox-Vi-update}) in proximal RADMM (by iteration equation (\ref{eq:2s-proxgra-Vi-update}) in proximal linearized RADMM).}
Update $\V_{N-1}^{k+1}$ by iteration equation (\ref{eq:2s-Vn-1-update}).\\
Update $\V_{N}^{k+1}$ by iteration equation (\ref{eq:2s-Vn-update}).\\
Update $\Lam^{k+1}$ by iteration equation (\ref{eq:2s-lambda-update}).}
\Output{$\{\W_{i}^K\}_{i=1}^{N}$.}
\end{algorithm}}

\subsection{Convergence (Rate) of Two-Splitting RADMMs}\label{sec:convergence-of-2-splitting-ADMM}
In this section, we establish convergence and analyze the convergence rates of both the iteration points and objective function values for the serial and parallel two-splitting proximal RADMM algorithms presented in Algorithms \ref{alg:2sADMM_s} and \ref{2sADMMp_prox}.

\begin{assumption}[Boundness]\label{assum:assumption2s2}
Parameters of the two-splitting proximal RADMM satisfy:

\noindent$\bullet$ $\beta>1$,

\noindent $\bullet$ $\underline{\omega}_i\le\omega_i^k\le \overline{\omega}_i$ with some $0<\underline{\omega}_i\leq\overline{\omega}_i, i\in[N-1]$,

\noindent $\bullet$ $\underline{\nu}_i\le\nu_i^k\le\overline{\nu}_i$ with some $0<\underline{\nu}_i\leq\overline{\nu}_i, i\in[N-1]$.
\end{assumption}

\noindent Under Assumption \ref{assum:assumption2s2}, we have \Cref{prop:2s_prox_theorem1} and \Cref{prop:2s_prox_theorem2}.
\begin{proposition}[Proof in Appendix \ref{app:1}]\label{prop:2s_prox_theorem1}
$\mathcal{L}_{\beta}^{2s}(\Psi_{2s}^{k+1})\le\mathcal{L}_{\beta}^{2s}(\Psi_{2s}^{k})-c_1\Vert \Psi_{2s}^{k+1}-\Psi_{2s}^{k}\Vert^2$ with some $c_1>0$.
\end{proposition}
\begin{proposition}[Proof in Appendix \ref{app:2}]\label{prop:2s_prox_theorem2}
$\Vert\nabla \mathcal{L}_{\beta}^{2s}(\Psi_{2s}^{k+1})\Vert\le c_2\Vert\Psi_{2s}^{k+1}-\Psi_{2s}^{k}\Vert$ with some $c_2 >0$.
\end{proposition}

\noindent Our two main convergence results on the two-splitting proximal RADMM are shown in \Cref{thm:thm_2s_prox_main_result1} and \Cref{thm:convergence_rate}.
\begin{theorem}[Convergence, proof in Appendix \ref{proof of theorems 1 and 2}]\label{thm:thm_2s_prox_main_result1}
$\Psi_{2s}^{k}\to \Psi_{2s}^{*}$ and $\mathcal{L}_{\beta}^{2s}(\Psi_{2s}^{k})\to \mathcal{L}_{\beta}^{2s}(\Psi_{2s}^{*})$. $\Psi_{2s}^{*}$ is a KKT point of (\ref{mod:2SARTM}).
\end{theorem}

\noindent \Cref{thm:thm_2s_prox_main_result1} guarantees the convergence of both iteration point and objective function value sequences. Specifically, the limit of the iteration is a KKT point.
\begin{theorem}[Convergence Rate, proof in Appendix \ref{proof of theorems 1 and 2}]\label{thm:convergence_rate}
Let $\theta$ be the KL exponent of $\mathcal{L}_{\beta}^{2s}$ at $\Psi_{2s}^*$.\\
\emph{1. (R-linear)} If $\theta=\frac{1}{2}$, then there exist $k_{0}\in\mathbb{N}$, $\eta\in(0, 1)$ and $c>0$ such that for each $k\geq k_{0}$, we have

$\bullet$ $\Vert \Psi_{2s}^{k}-\Psi_{2s}^{*}\Vert\leq c\eta^{k-k_{0}+1}$,

\vspace{1mm}
$\bullet$ $\mathcal{L}_{\beta}^{2s}(\Psi_{2s}^{k})-\mathcal{L}_{\beta}^{2s}(\Psi_{2s}^{*})\leq c\eta^{k-k_{0}+1}$.\\
\emph{2. (R-sublinear)} If $\theta\in(\frac{1}{2}, 1)$, then there exist $k_{0}\in\mathbb{N}$ and $c>0$ such that for each $k\geq k_{0}$,

$\bullet$ $\Vert \Psi_{2s}^{k}-\Psi_{2s}^{*}\Vert\leq c(k-k_{0}+1)^{\frac{1-\theta}{1-2\theta}}$,

\vspace{1mm}
$\bullet$ $\mathcal{L}_{\beta}^{2s}(\Psi_{2s}^{k})-\mathcal{L}_{\beta}^{2s}(\Psi_{2s}^{*})\leq c(k-k_{0}+1)^{-\frac{1}{2\theta-1}}$.
\end{theorem}

\vspace{-1mm}
\noindent Theorem \ref{thm:convergence_rate} implies the \emph{R-linear} and $\mathcal{O}(k^{\frac{1-\theta}{1-2\theta}})$ ($\mathcal{O}(k^{-\frac{1}{2\theta - 1}})$) \emph{R-sublinear} convergence rate of the iteration point (objective function value) sequences generated by the proximal two-splitting RADMM training algorithms.

\section{Three-splitting RADMMs}\label{sec:3-splitting-ADMM}
In this section, to facilitate the efficient training of deeper residual networks, we design and discuss scalable three-splitting proximal (linearized) RADMM training algorithms.

\subsection{Three-Splitting Proximal (Linearized) RADMMs}
Clearly, the three-splitting approximation (\ref{mod:ResNet-training-model5}) with MSE loss, ridge regression penalty, and identity mapping $h_i$, $i\in[N-1]$ is as below.
\vspace{-2mm}
{\footnotesize\begin{align}
\footnotesize
~\min_{\W, \U, \V}~& \mathcal{J}\big(\W, \V_N\big)+\frac{\mu}{2}\sum_{i=1}^{N-1}\Vert \V_{i-1}+\sigma_i(\U_{i})-\V_{i}\Vert^{2}\nonumber\\
\mbox{s.t.}~
& \U_{i}=\W_{i}\V_{i-1}, i\in[N-1],\nonumber\\
& \V_{N}=\W_{N}\V_{N-1},\label{mod:3SARTM}
\end{align}}
\vspace{-5mm}

\noindent where $\U_i=(\ui_i^1, \ldots, \ui_i^n)$ for $i\in [N-1]$, $\U := (\veccommand(\U_1),$\\$\ldots,\veccommand(\U_{N-1}))$, whose augmented Lagrangian function is
\vspace{-1.5mm}
{\footnotesize\begin{align}
\begin{split}\label{eq:3SARTM-ALF}
\footnotesize
&\mathcal{L}_{\beta}^{3s}(\Psi_{3s})\coloneqq \mathcal{J}\big(\W, \V_N\big)+\frac{\mu}{2}\sum_{i=1}^{N-1}\Vert \V_{i-1}+\sigma_i(\U_{i})-\V_{i}\Vert^{2}\\
&+\sum_{i=1}^{N-1}\big(\langle\Lam_{i}, \W_{i}\V_{i-1}-\U_{i}\rangle+\frac{\beta_{i}}{2}\Vert \W_{i}\V_{i-1}-\U_{i}\Vert^{2}\big)\\
&+\langle\Lam_{N}, \W_{N}\V_{N-1}-\V_{N}\rangle+\frac{\beta_{N}}{2}\Vert \W_{N}\V_{N-1}-\V_{N}\Vert^{2},
\end{split}
\end{align}}

\noindent in which $\Psi_{3s} := (\{\veccommand(\W_i)\}_{i=1}^N, \{\veccommand(\V_i)\}_{i=1}^N, \{\veccommand(\Lam_i)\}_{i=1}^N, $\\ $\{\veccommand(\U_i)\}_{i=1}^{N-1})$, $\{\Lam_i\}_{i=1}^{N-1}\subseteq\mathbb{R}^{d\times n}$ and $\Lam_N\in\mathbb{R}^{q\times n}$ are dual variables, parameters $\beta_i>0, i\in[N]$. Update subproblems of each block variable based on (\ref{eq:3SARTM-ALF}) in a Gauss-Seidel manner are listed in \Cref{table:update 3-splitting}. After that, the pseudocode of three-splitting proximal (linearized) RADMM training algorithms are given in Algorithm \ref{alg:3sADMM_s}.
{\footnotesize\begin{table}[h]
\caption{Updates of block variables in three-splitting RADMMs.}
\renewcommand\arraystretch{0.01}
\footnotesize
\centering\label{table:update 3-splitting}
    \begin{tabularx}{0.489\textwidth}{p{0.34in}| p{2.8in}}
        \hline
 Variables &~~~~~~~~~~~~~~~~~~~~~~~~~~~~~~~Updates\\
        \hline
\begin{align*}\W_N^{k+1}\end{align*} & \begin{align}
\label{eq:3s-Wn-update}
(\beta_{N}\V_{N}^{k}-\Lam_{N}^{k}){\V_{N-1}^{k~\mathrm{T}}}(\lambda \I+\beta_{N}\V_{N-1}^{k}{\V_{N-1}^{k~\mathrm{T}}})^{-1}
\end{align}\\
\hline
\begin{align*}\W_i^{k+1}\end{align*} & \begin{align}
\label{eq:3s-Wi-update}
(\beta_{i}\U_{i}^{k}-\Lam_{i}^{k})\V_{i-1}^{k~\mathrm{T}}
(\lambda \I+\beta_{i}\V_{i-1}^{k}\V_{i-1}^{k~\mathrm{T}})^{-1}~~~~~~~~
\end{align}\\
\hline
    \multirow{1200}{=}{\begin{align*}\\ \\ \\ \\\U_i^{k+1}\end{align*}}
 &   \begin{align}\label{eq:3s-prox-Ui-update}
       \text{P:}~~\argmin\limits_{\U_{i}}\big\{\frac{\mu}{2}\Vert \V_{i-1}^{k+1}+\sigma_i(\U_{i})-\V_{i}^{k}\Vert^{2}~~~~~~~~~~~\nonumber\\
+\frac{\beta_{i}}{2}\big\Vert \U_{i}-\W_{i}^{k+1}\V_{i-1}^{k+1}-\frac{1}{\beta_{i}}\Lam_{i}^{k}\big\Vert^{2}~~~~~~~~~~~~~\nonumber\\
+\frac{\omega_{i}^{k}}{2}\Vert \U_{i}-\U_{i}^{k}\Vert^{2}\big\}~~~~~~~~~~~~~~~~~~~~~~~~~~~~~~~~~~
                        \end{align}\\
        \cline{2-2}
  & \begin{align}
       \label{eq:3s-proxgra-Ui-update}
\text{PL:}~~-\frac{\mu}{\tau_{i}^{k}+\beta_{i}}\big(\V_{i-1}^{k+1}+\sigma_i(\U_{i}^{k})-\V_{i}^{k}\big)\odot~~~~~~~~~\nonumber\\
\sigma_i^{'}(\U_{i}^{k})+\frac{\tau_{i}^{k}}{\tau_{i}^{k}+\beta_{i}} \U_{i}^{k}+\frac{1}{\tau_{i}^{k}+\beta_{i}}\Lam_{i}^{k}~~~~~~~~~\nonumber\\
+\frac{\beta_{i}}{\tau_{i}^{k}+\beta_{i}}\W_{i}^{k+1}\V_{i-1}^{k+1}~~~~~~~~~~~~~~~~~~~~~~~~~~~
\end{align} \\
\hline
\begin{align*}\\ \\ \V_i^{k+1}\end{align*} & \begin{align}
\label{eq:3s-Vi-update}
\mu\big(2\mu \I+\beta_{i+1}\W_{i+1}^{k+1~\mathrm{T}}\W_{i+1}^{k+1}\big)^{-1}
\big(\V_{i-1}^{k+1}+\V_{i+1}^{k}\nonumber\\
-\sigma_{i+1}(\U_{i+1}^{k})+\sigma_i(\U_{i}^{k+1})\big)
+\beta_{i+1}\big(2\mu \I+~~~~~~~~\nonumber
\\\beta_{i+1}\W_{i+1}^{k+1~\mathrm{T}}\W_{i+1}^{k+1}\big)^{-1}
\W_{i+1}^{k+1~\mathrm{T}}\U_{i+1}^{k}~~~~~~~~~~~~\nonumber\\
-\big(2\mu \I+\beta_{i+1}\W_{i+1}^{k+1~\mathrm{T}}
\W_{i+1}^{k+1}\big)^{-1}
\W_{i+1}^{k+1~\mathrm{T}}\Lam_{i+1}^{k}
\end{align}\\
  \hline
\begin{align*}\\ \\ \V_{N-1}^{k+1}\end{align*} & \begin{align}
\label{eq:3s-Vn-1-update}
\mu\big(\mu \I+\beta_{N}\W_{N}^{k+1~\mathrm{T}}\W_{N}^{k+1})^{-1}(\sigma_{N-1}(\U_{N-1}^{k+1})\nonumber\\
+\V_{N-2}^{k+1}\big)
+\beta_{N}\big(\mu \I+\beta_{N}\W_{N}^{k+1~\mathrm{T}}\W_{N}^{k+1}\big)^{-1}~\nonumber\\
\W_{N}^{k+1~\mathrm{T}}\V_{N}^{k}-\big(\mu \I+\beta_{N}\W_{N}^{k+1~\mathrm{T}}
\W_{N}^{k+1}\big)^{-1}~\nonumber\\
\W_{N}^{k+1~\mathrm{T}}\Lam_{N}^{k}~~~~~~~~~~~~~~~~~~~~~~~~~~~~~~~~~~~~~~~~~~
\end{align}\\
\hline
\begin{align*}\V_{N}^{k+1}\end{align*} & \begin{align}\label{eq:3s-Vn-update}\frac{1}{1+\beta_{N}}\big(\Y+\beta_{N}\W_{N}^{k+1}\V_{N-1}^{k+1}+\Lam_{N}^{k}\big)~~~~~~~~~~\end{align}\\
\hline\\
\begin{align*}\Lam_i^{k+1}\end{align*} &
\begin{align}\label{eq:3s-lambdai-update}\Lam_{i}^{k}+\beta_{i}\big(\W_{i}^{k+1}\V_{i-1}^{k+1}-\U_{i}^{k+1}\big)~~~~~~~~~~~~~~~~~\end{align}
\\
\hline\\
\begin{align*}\Lam_N^{k+1}\end{align*} &
\begin{align}\label{eq:3s-lambdan-update}\Lam_{N}^{k}+\beta_{N}\big(\W_{N}^{k+1}\V_{N-1}^{k+1}-\V_{N}^{k+1}\big)~~~~~~~~~~~~~~\end{align}
\\
\hline
    \end{tabularx}
\end{table}}

{\footnotesize\begin{algorithm}[h]\label{alg:3sADMM_s}
\footnotesize
\SetKwInput{Initialization}{Initialization}
\SetKw{Return}{return}
\SetKwFor{ForParallel}{parallel\_for}{do}{end}
\SetKwFunction{Range}{range}
\SetKwInput{Input}{Input}
\SetKwInput{Output}{Output}

\caption{Three-Splitting RADMMs}
\Input{$\X, \Y, K, \lambda, \mu, \{\beta_{i}\}_{i=1}^N$, $\{\omega_i^k\}_{i=1}^{N-1}$, and $\{\W_{i}^{0}\}_{i=1}^N$.}
$\V_{0}^{k}\gets \X$, $\U_{i}^{0}\gets \W_{i}^{0}\V_{i-1}^{0}$, $\V_{i}^{0}\gets \V_{i-1}^{0}+\sigma_i(\U_{i}^{0})$, $i\in[N-1]$, $\V_{N}^{0}\gets \W_{N}^{0}\V_{N-1}^{0}$, and $\Lam_{i}^{0}\gets \boldsymbol{0}$, $i\in[N]$.\\
\For{$k\gets0$~{\rm\textbf{to}}~$K-1$}{
Update $\W_N^{k+1}$ by iteration equation (\ref{eq:3s-Wn-update}).\\
\For{$i\gets N-1$~{\rm\textbf{to}}~$1$}{
Update $\W_{i}^{k+1}$ by iteration equation (\ref{eq:3s-Wi-update}).}
\For{$i\gets1$~{\rm\textbf{to}}~$N-2$}{
Update $\U_{i}^{k+1}$ by solving (\ref{eq:3s-prox-Ui-update}) in proximal RADMM (by iteration equation (\ref{eq:3s-proxgra-Ui-update}) in proximal linearized RADMM).\\
Update $\V_{i}^{k+1}$ by iteration equation (\ref{eq:3s-Vi-update}).}
Update $\U_{N-1}^{k+1}$ by solving (\ref{eq:3s-prox-Ui-update}) in proximal RADMM (by iteration equation (\ref{eq:3s-proxgra-Ui-update}) in proximal linearized RADMM).\\
Update $\V_{N-1}^{k+1}$ by iteration equation (\ref{eq:3s-Vn-1-update}).\\
Update $\V_{N}^{k+1}$ by iteration equation (\ref{eq:3s-Vn-update}).\\
\For{$i\gets1$~{\rm\textbf{to}}~$N-1$}{
Update $\Lam_i^{k+1}$ by iteration equation (\ref{eq:3s-lambdai-update}).}
Update $\Lam_N^{k+1}$ by iteration equation (\ref{eq:3s-lambdan-update}).\\}
\Output{$\{\W_{i}^K\}_{i=1}^N$.}
\end{algorithm}}

\subsection{Convergence (Rate) of Three-Splitting RADMMs}\label{sec:convergence-of-3-Splitting-ADMM}
In this section, we establish convergence and derive convergence rate estimates for both the iteration points and the objective function values of the serial and parallel three-splitting proximal RADMMs. The challenges encountered in the theoretical analysis and our proof outline are discussed in \Cref{subsubsection:Regularized Augmented Lagrangian} and \Cref{subsubsection:connection}, respectively, with the principal results presented in \Cref{subsubsec:convergence results}.

\subsubsection{Regularized Augmented Lagrangian}\label{subsubsection:Regularized Augmented Lagrangian} Unfortunately, different from the $\mathcal{L}_\beta^{2s}$, it is difficult to realize a sufficient descent for $\mathcal{L}_\beta^{3s}$ directly. To deal with this issue, we first construct a regularization $\mathcal{L}_{R}^{3s}$ for the augmented Lagrangian function:
\vspace{-1.5mm}
{\small\begin{align*}
\begin{split}
\small
\mathcal{L}_{R}^{3s}(\widetilde\Psi_{3s})\coloneqq&\mathcal{L}_{\beta}^{3s}(\Psi_{3s})+\sum_{i=1}^{N-1}\theta_{i}\Vert \U_{i}-\widetilde\U_{i}\Vert^{2}+\sum_{i=1}^{N-1}\eta_{i}\Vert \V_{i}-\widetilde\V_{i}\Vert^{2},
\end{split}
\end{align*}}
\vspace{-3.5mm}

\noindent where
$\widetilde\Psi_{3s}\coloneqq(\Psi_{3s}, \{\veccommand(\widetilde\U_{i})\}_{i=1}^{N-1}, \{\veccommand(\widetilde\V_{i})\}_{i=1}^{N-1})$, $\theta_i\coloneqq\frac{4(\underline{\omega}_i)^2}{\beta_i}+\frac{\underline{\omega}_i}{4},~ \eta_i\coloneqq\frac{4\mu^2\psi_1^2}{\beta_i}+\frac{\mu}{4}$, $i\in[N-1]$. We define the iteration $\widetilde\U_{i}^{k}\coloneqq \U_{i}^{k-1}$, $\widetilde\V_{i}^{k}\coloneqq \V_{i}^{k-1}$, $i\in[N-1]$. Similarly with the discussion on two-splitting RADMMs in \Cref{sec:convergence-of-2-splitting-ADMM}, we can prove that (see Lemma \ref{lem:1})
\vspace{-1mm}
{\small\begin{align*}
&\widetilde\Psi_{3s}^k\to\widetilde\Psi_{3s}^*,~ \mathcal{L}_{R}^{3s}(\widetilde\Psi_{3s}^k)\to \mathcal{L}_{R}^{3s}(\widetilde\Psi_{3s}^*),~ \nabla \mathcal{L}_R^{3s}(\widetilde\Psi_{3s}^*) = \boldsymbol{0};\\
& \Vert\widetilde\Psi_{3s}^k-\widetilde\Psi_{3s}^*\Vert = \mathcal{O}(\eta^k) ~(\mathcal{O}(k^{\frac{1-\theta}{1-2\theta}})),\\ &\mathcal{L}_{R}^{3s}(\widetilde\Psi_{3s}^{k})-\mathcal{L}_{R}^{3s}(\widetilde\Psi_{3s}^{*})=\mathcal{O}(\eta^k) ~(\mathcal{O}(k^{-\frac{1}{2\theta-1}})).
\end{align*}}

\subsubsection{Connections between $\mathcal{L}_{\beta}^{3s}(\Psi_{3s})$ and $\mathcal{L}_R^{3s}(\widetilde\Psi_{3s})$}\label{subsubsection:connection}
We can prove that
\vspace{-2mm}
{\small\begin{align}
\begin{split}
&\Vert \Psi_{3s}^k-\Psi_{3s}^*\Vert\le\Vert \widetilde\Psi_{3s}^k-\widetilde\Psi_{3s}^*\Vert,\\
&\mathcal{L}_{\beta}^{3s}(\Psi_{3s}^{k})-\mathcal{L}_{\beta}^{3s}(\Psi_{3s}^*)\le\mathcal{L}_{R}^{3s}(\widetilde\Psi_{3s}^{k})-\mathcal{L}_{R}^{3s}(\widetilde\Psi_{3s}^*).
\end{split}
\end{align}}
\vspace{-1.5mm}

\noindent Then the convergence of $\{\widetilde\Psi_{3s}^k\}$ and $\{\mathcal{L}_{R}^{3s}(\widetilde\Psi_{3s}^k)\}$ imply that of $\{\Psi_{3s}^k\}$ and $\{\mathcal{L}_{\beta}^{3s}(\Psi_{3s}^k)\}$. In addition, we have $\nabla \mathcal{L}_R^{3s}(\widetilde\Psi_{3s}^*) = \boldsymbol{0}$, and for each $i\in[N-1]$,
\vspace{-1mm}
{\small\begin{align}
\footnotesize
\frac{\partial \mathcal{L}_{R}^{3s}}{\partial \U_{i}}(\widetilde\Psi_{3s}^{*})
=\frac{\partial \mathcal{L}_{\beta}^{3s}}{\partial \U_{i}}(\Psi_{3s}^*),~~
\frac{\partial \mathcal{L}_{R}^{3s}}{\partial \V_{i}}(\widetilde\Psi_{3s}^{*})
=\frac{\partial \mathcal{L}_{\beta}^{3s}}{\partial \V_{i}}(\Psi_{3s}^*).
\end{align}}

\begin{figure}[h]
  \centering
  \includegraphics[width=1\linewidth]{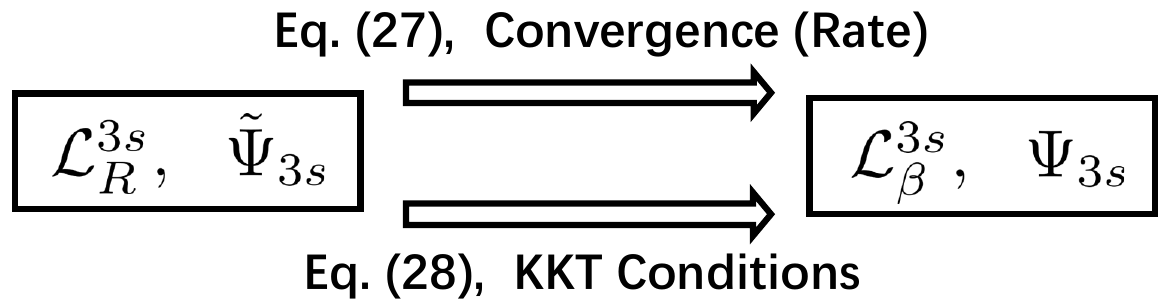}
\caption{Relationship between $\mathcal{L}_{\beta}^{3s}(\Psi_{3s})$ and $\mathcal{L}_R^{3s}(\widetilde\Psi_{3s})$.}
		\label{fig:relationship-auxiliary-function}
\end{figure}

\noindent Thus it can be proved that the limit $\Psi_{3s}^*$ is a KKT point (see \Cref{thm:3}). Furthermore, the same convergence rate can also be guaranteed (see \Cref{thm:3s_prox_theorem6}).
Relationship between $\mathcal{L}_{\beta}^{3s}(\Psi_{3s})$ and $\mathcal{L}_R^{3s}(\widetilde\Psi_{3s})$ is shown in \Cref{fig:relationship-auxiliary-function}.
\vspace{-3mm}
\vspace{2mm}
\subsubsection{Convergence results}\label{subsubsec:convergence results}
Boundness assumption is first given in \Cref{assum:3s_prox_assumption2}.
\begin{assumption}[Boundness]\label{assum:3s_prox_assumption2}
Parameters of the three-splitting proximal RADMM satisfy:

\noindent $\bullet$ $\beta_i > \underline{\beta}_i$ with some $\underline{\beta}_i, i \in [N]$,

\vspace{1mm}
\noindent $\bullet$
$\underline{\omega}_{i}<\omega_{i}^{k}\leq\omega_{i}^{k+1}<\overline{\omega}_i$ with some $\underline{\omega}_i<\overline{\omega}_i$, $i\in[N-1]$.
\end{assumption}
\noindent See Appendix \ref{subsec:values} for the values of $\{\underline{\beta}_i\}_{i=1}^N$, $\{\underline{\omega}_i\}_{i=1}^{N-1}$, and $\{\overline{\omega}_i\}_{i=1}^{N-1}$. Moreover, we assume the upper boundedness $\Vert \V_{i}^{k}\Vert\leq\overline{\mathcal{V}}_{i}$ with some $\overline{\mathcal{V}}_{i}>0$,
$i\in[N-1]$. Under these assumptions, we first have \Cref{lem:1} about $\mathcal{L}_{R}^{3s}$ and $\widetilde\Psi_{3s}$.
\begin{lemma}[Proof in Appendix \ref{subsection-theorem6-theorem7}]\label{lem:1}~\\
$\widetilde\Psi_{3s}^{k}\to \widetilde\Psi_{3s}^{*}$, $\mathcal{L}_{R}^{3s}(\widetilde\Psi_{3s}^{k})\to \mathcal{L}_{R}^{3s}(\widetilde\Psi_{3s}^{*})$.

\noindent \emph{1.} If the KL exponent $\theta$ of $\mathcal{L}_R^{3s}$ at $\widetilde\Psi_{3s}^*$ is equal to $\frac{1}{2}$, then there exist $k_{0}\in\mathbb{N}$, $\eta\in(0, 1)$ and $c>0$ such that for each $k\geq k_{0}$, we have
$\Vert \widetilde\Psi_{3s}^{k}-\widetilde\Psi_{3s}^{*}\Vert\leq c\eta^{k-k_{0}+1}$, and $\mathcal{L}_R^{3s}(\widetilde\Psi_{3s}^{k})-\mathcal{L}_R^{3s}(\widetilde\Psi_{3s}^{*})\leq c\eta^{k-k_{0}+1}$.\\
\emph{2.} If $\theta\in(\frac{1}{2}, 1)$, then there exist $k_{0}\in\mathbb{N}$ and $c>0$ such that for each $k\geq k_{0}$, we have
$\Vert \widetilde\Psi_{3s}^{k}-\widetilde\Psi_{3s}^{*}\Vert\leq c(k-k_{0}+1)^{\frac{1-\theta}{1-2\theta}}$, and $\mathcal{L}_R^{3s}(\widetilde\Psi_{3s}^{k})-\mathcal{L}_R^{3s}(\widetilde\Psi_{3s}^{*})\leq c(k-k_{0}+1)^{-\frac{1}{2\theta-1}}$.
\end{lemma}
Our main convergence results of the three-splitting proximal RADMMs are shown in \Cref{thm:3} and \Cref{thm:3s_prox_theorem6}.
\begin{theorem}[Convergence, proof in Appendix \ref{subsection-theorem6-theorem7}]\label{thm:3}
$\Psi_{3s}^{k}\to \Psi_{3s}^{*}$ and $\mathcal{L}_{\beta}^{3s}(\Psi_{3s}^{k})\to \mathcal{L}_{\beta}^{3s}(\Psi_{3s}^{*})$. $\Psi_{3s}^{*}$ is a KKT point of (\ref{mod:3SARTM}).
\end{theorem}
\begin{theorem}[Convergence rate, proof in Appendix \ref{subsection-theorem6-theorem7}]\label{thm:3s_prox_theorem6}
Convergence rate of the three-splitting proximal RADMM:\\
\emph{1. (R-linear)} If the KL exponent $\theta=\frac{1}{2}$, then there exist $k_{0}\in\mathbb{N}$, $\eta\in(0, 1)$ and $c>0$ such that for each $k\geq k_{0}$, we have

$\bullet$ $\Vert \Psi_{3s}^{k}-\Psi_{3s}^{*}\Vert\leq c\eta^{k-k_{0}+1}$,

\vspace{1mm}
$\bullet$ $\mathcal{L}_{\beta}^{3s}(\Psi_{3s}^{k})-\mathcal{L}_{\beta}^{3s}(\Psi_{3s}^{*})\leq c\eta^{k-k_{0}+1}$.\\
\emph{2. (R-sublinear)} If $\theta\in(\frac{1}{2}, 1)$, then there exist $k_{0}\in\mathbb{N}$ and$c>0$ such that for each $k\geq k_{0}$,

$\bullet$ $\Vert \Psi_{3s}^{k}-\Psi_{3s}^{*}\Vert\leq c(k-k_{0}+1)^{\frac{1-\theta}{1-2\theta}}$,

\vspace{1mm}
$\bullet$ $\mathcal{L}_{\beta}^{3s}(\Psi_{3s}^{k})-\mathcal{L}_{\beta}^{3s}(\Psi_{3s}^{*})\leq c(k-k_{0}+1)^{-\frac{1}{2\theta-1}}$.
\end{theorem}

\section{Parallel and distributed algorithms}\label{sec:advantages-of-parallel-implementation}
In this section, we develop the parallel/distributed RADMM training algorithms and theoretically demonstrate their advantages in terms of reduced time complexity and lower per-node runtime memory requirements.

\subsection{Parallel RADMMs}\label{subsec:parallel admms}
\begin{figure}[h]
  \centering
  \includegraphics[width=1\linewidth]{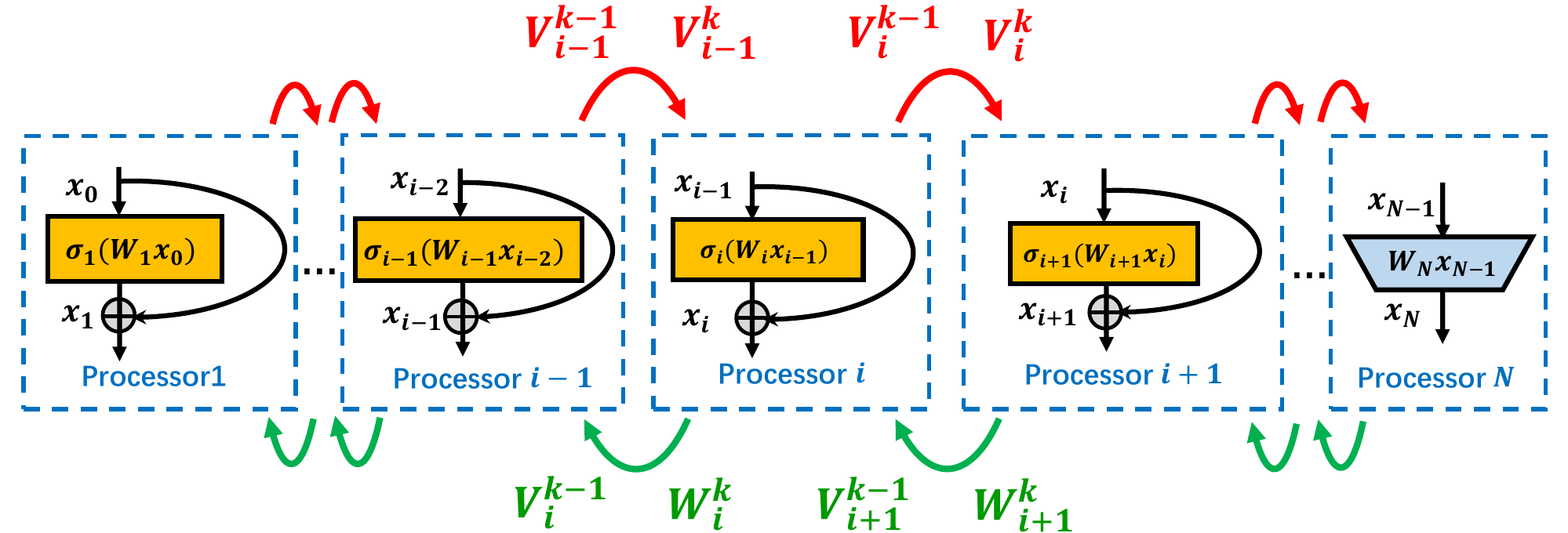}
\caption{PRU employed in parallel RADMMs.}
		\label{fig:2s-p}
\end{figure}

Note that for each residual block in the network, updating its block variables requires only the block variables from its immediately adjacent block(s), which presents an opportunity for model parallelism. We refer to this scheme as the parallel regional update (PRU), where ``regional'' highlights this localized update characteristic. As illustrated in \Cref{fig:2s-p}, the arrows indicate the communication between adjacent processors (or blocks). In contrast, BP-based methods necessitate that the gradient be sequentially propagated through all blocks, thereby limiting scalability. With PRU, we can assign
$N$ parallel processors, each responsible for updating the block variables of a single residual block.
\vspace{-0.5mm}
Based on this observation, we propose parallel RADMM training algorithms derived from our serial version. In our approach, a processor initiates the update of its assigned block as soon as the necessary block variables from the adjacent block(s) become available. Note that the
$N$ processors operate asynchronously, as a processor must wait to retrieve a variable from another until it has been updated. However, such waiting does not fully serialize the updates, rather, it produces a pipelined update pattern that improves the time complexity relative to the serial version (see \Cref{subsec:lower-time-complexity}). The implementation strategy and details are provided in \Cref{subsec:parallel-implementation}.
{\footnotesize\begin{algorithm}[h]\label{2sADMMp_prox}
\footnotesize
\caption{Parallel Two-Splitting RADMMs}
\SetKwInput{Initialization}{Initialization}
\SetKw{Return}{return}
\SetKwFor{ForParallel}{parallel\_for}{do}{end}
\SetKwFunction{Range}{range}
\SetKwInput{Input}{Input}
\SetKwInput{Output}{Output}

\Input{$\X$, $\Y$, $K$, $\lambda$, $\mu$, $\beta$, $\{\omega_i^k\}_{i=1}^{N-1}$, $\{\nu_i^k\}_{i=1}^{N-1}$, and $\{\W_{i}^{0}\}_{i=1}^N$.}
$\V_{0}^{k}\gets \X$, $\V_{i}^{0}\gets \V_{i-1}^{0}+\sigma_i(\W_{i}^{0}\V_{i-1}^{0})$, $i\in[N-1]$,
$\V_{N}^{0}\gets \W_{N}^{0}\V_{N-1}^{0}$ and $\Lam^{0}\gets \boldsymbol{0}$.\\
\ForParallel{$i\in[N]$}{
    \For{$k\gets0$~{\rm\textbf{to}}~$K-1$}{
        \eIf{$i < N$}{
            Update $\W_{i}^{k+1}$ by solving (\ref{eq:2s-prox-Wi-update}) in proximal
RADMM (by iteration equation (\ref{eq:2s-proxgra-Wi-update}) in proximal linearized RADMM).
        }{
        Update $\W_{N}^{k+1}$ by iteration equation (\ref{eq:2s-Wn-update}).
        }
        \lIf{$i < N$}{Retrieve necessary block variables from processor $i+1$}
        \lIf{$i > 1$}{Retrieve necessary block variables from processor $i-1$}
        \uIf{$i< N-1$}{
               Update $\V_{i}^{k+1}$ by solving (\ref{eq:2s-prox-Vi-update}) in proximal RADMM (by iteration equation (\ref{eq:2s-proxgra-Vi-update}) in proximal linearized RADMM).
         }
        \uElseIf{$i=N-1$}{
        Update $\V_{N-1}^{k+1}$ by iteration equation (\ref{eq:3s-Vn-1-update}).
        }
        \Else{
        Update $\V_{N}^{k+1}$ by iteration equation (\ref{eq:3s-Vn-update}).
        }
        \If{$i = N$}{
            Update $\Lam^{k+1}$ by iteration equation (\ref{eq:2s-lambda-update}).
        }
    }
}
Synchronize all processors.\\
\Output{$\{\W_{i}^K\}_{i=1}^{N}$.}
\end{algorithm}}
\subsection{Lower Time Complexity}\label{subsec:lower-time-complexity}
To demonstrate the reduced time complexity of our parallel version, we compare the time complexities of the serial and parallel proximal linearized RADMMs. For illustration, \Cref{fig:pipeline} depicts a pipelined update pattern for the parallel RADMMs. In this figure, the horizontal axis represents runtime, where each unit of time corresponds to the update duration for a single block variable. Each row represents the update sequence for one block variable, and in the leftmost column, block variables grouped within the same box are updated by the same processor. The number in each block indicates the epoch during which the corresponding variable is updated.
\begin{figure}[h]
  \centering
  \includegraphics[width=1\linewidth]{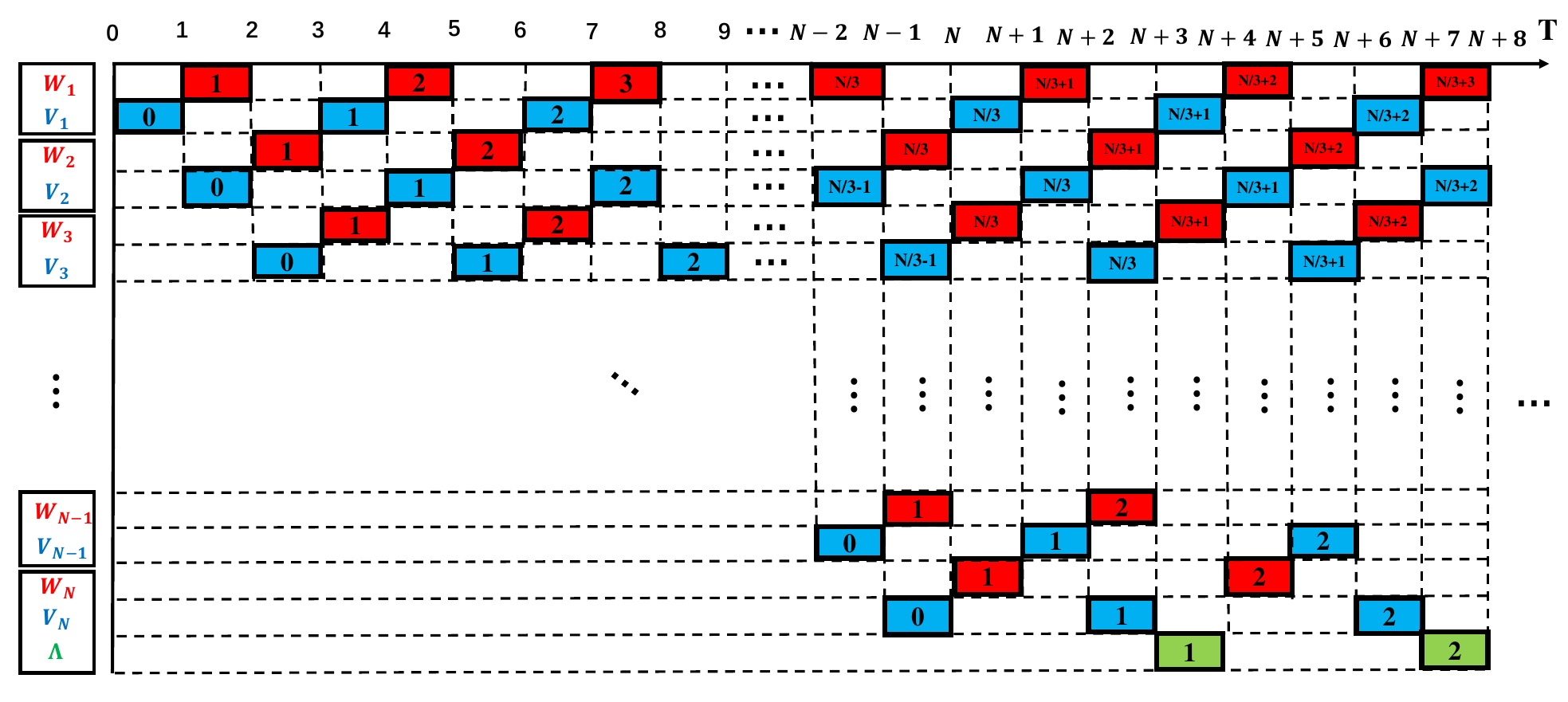}
\caption{Pipelined update pattern of the parallel two-splitting RADMMs.}
		\label{fig:pipeline}
\end{figure}

Let $T_{\scriptscriptstyle \rm mul}(n)$ denote the computational complexity of multiplying two $n\times n$ matrices, and let $T_{\scriptscriptstyle \rm comm}(K, N, d, q, n)$ denote the communication cost incurred by processors during $K$ updates. Here, basic operations include addition, subtraction, multiplication, division, and the evaluation of activation functions. Let $\tau = \max\{d, q, n\}$. According to recent studies on the time complexity of matrix multiplication \cite{Williams2024},
the cost of square matrix multiplication constitutes the primary bottleneck in updating block variables. Therefore, the time complexity for each block variable update is $\mathcal{O}(T_{\scriptscriptstyle \rm mul}(\tau))$.
\subsubsection{Serial RADMMs}
The time complexities of serial proximal RADMM training algorithms are both $\mathcal{O}(KNT_{\scriptscriptstyle \rm mul}(\tau))$.

\subsubsection{Parallel RADMMs} Based on the pipeline shown in Figure \ref{fig:pipeline}, the time complexities of the parallel RADMMs are
\vspace{-1.5mm}
\begin{align*}\mathcal{O}(\max\{K, N\}T_{\scriptscriptstyle \rm mul}(\tau))+\mathcal{O}(T_{\scriptscriptstyle \rm comm}(K, N, d, q, n)).\end{align*}
\vspace{-6mm}

\noindent If the communication cost is relatively small, i.e.,
\vspace{-1.5mm}
\begin{align*}\max\{K, N\}T_{\scriptscriptstyle \rm mul}(\tau)=\mathcal{O}(T_{\scriptscriptstyle \rm comm}(K, N, d, q, n)),\end{align*}
\vspace{-6mm}

\noindent our parallelization via PRU can reduce the time complexity from $\mathcal{O}(KNT_{\scriptscriptstyle \rm mul}(\tau))$ to $\mathcal{O}(\max\{K, N\}T_{\scriptscriptstyle \rm mul}(\tau))$.

\subsection{Less Runtime Memory Requirement}\label{subsec:lower-storage-requirement}
In this section, we theoretically demonstrate the scalability of our distribution strategy. As noted in \Cref{sec:introduction}, BP-based methods do not exhibit this property. Under our approach, each node stores only the block variables corresponding to its assigned residual block, thereby substantially reducing the per-node memory footprint. Consequently, the distributed implementation of our parallel algorithms significantly alleviates memory constraints.
\subsubsection{Serial RADMMs}
For the serial algorithms, all variables are stored on a single processor, and only the iteration values from two consecutive steps need to be retained. Consequently, the runtime memory requirement for the serial RADMMs is $\mathcal{O}(N\max\{d, q\}\max\{d, n\})$.

\subsubsection{Distributed RADMMs}
In the distributed configuration, only the block variables allocated to each node are stored in its runtime memory. The per-node runtime memory requirements of the parallel RADMMs are summarized below, where processor
$i$ corresponds to the $i$th layer of the residual network:

\vspace{-0.5mm}
\begin{itemize}
\item Processors $1, \ldots, N-2$: $\mathcal{O}(d\max\{d,n\})$,
\item Processor $N-1$: $\mathcal{O}(\max\{d, q\}\max\{d, n\})$,
\item Processor $N$: $\mathcal{O}(\max\{q\max\{d, n\}, dn\})$.
\end{itemize}
\vspace{-1mm}
Therefore, we know that the distributed algorithm can reduce the per-node runtime memory requirement from \emph{cubic} to \emph{quadratic} complexity.

\section{EXPERIMENTS}\label{sec:experiments}
We compare our proximal linearized RADMMs with several well-known BP-based algorithms (SGD, SGDM, Adam) on the Wine Quality dataset\footnote{\url{https://archive.ics.uci.edu/dataset/186/wine+quality}} to demonstrate the fast and stable convergence, superior performance, and enhanced speed offered by our algorithms in \Cref{subsec:wine quality dataset}. Furthermore, the advantages of our parallel RADMMs are presented in Section \Cref{subsec:parallel-implementation}. Detailed experimental settings are provided in Appendix \ref{subsec:experimental settings}.

\vspace{-1mm}
\subsection{Wine Quality Dataset}\label{subsec:wine quality dataset}
In this section, we present experimental results on the Wine Quality dataset, demonstrating the fast and stable convergence, superior performance, and enhanced computational speed of our proximal linearized RADMM training algorithms.

\subsubsection{Fast and Stable Convergence}
We employ a 40-layer residual network to demonstrate the fast and stable convergence of our RADMM algorithms. Figure 5 presents the MSE for both training and testing losses for each algorithm after 600 iterations\footnote{The number of iterations in the experiments in this paper is equal to the number of updates divided by the number of batches in the train set.}. The results clearly indicate that our RADMMs converge rapidly with minimal fluctuations for networks utilizing both ReLU and sigmoid activations. Specifically, for the ReLU activation network, our RADMMs substantially outperform SGD, SGDM, and Adam by achieving lower MSE values and faster convergence. Similarly, for the sigmoid activation network, both the two- and three-splitting RADMM training algorithms attain lower MSE losses compared to SGD and SGDM, and perform comparably to Adam.
\begin{figure}[htb]
\begin{minipage}[htb]{.48\linewidth}
\centering\label{fig:40_layers_relu_WQ_test_MSE_loss}
\includegraphics[width=1\textwidth]{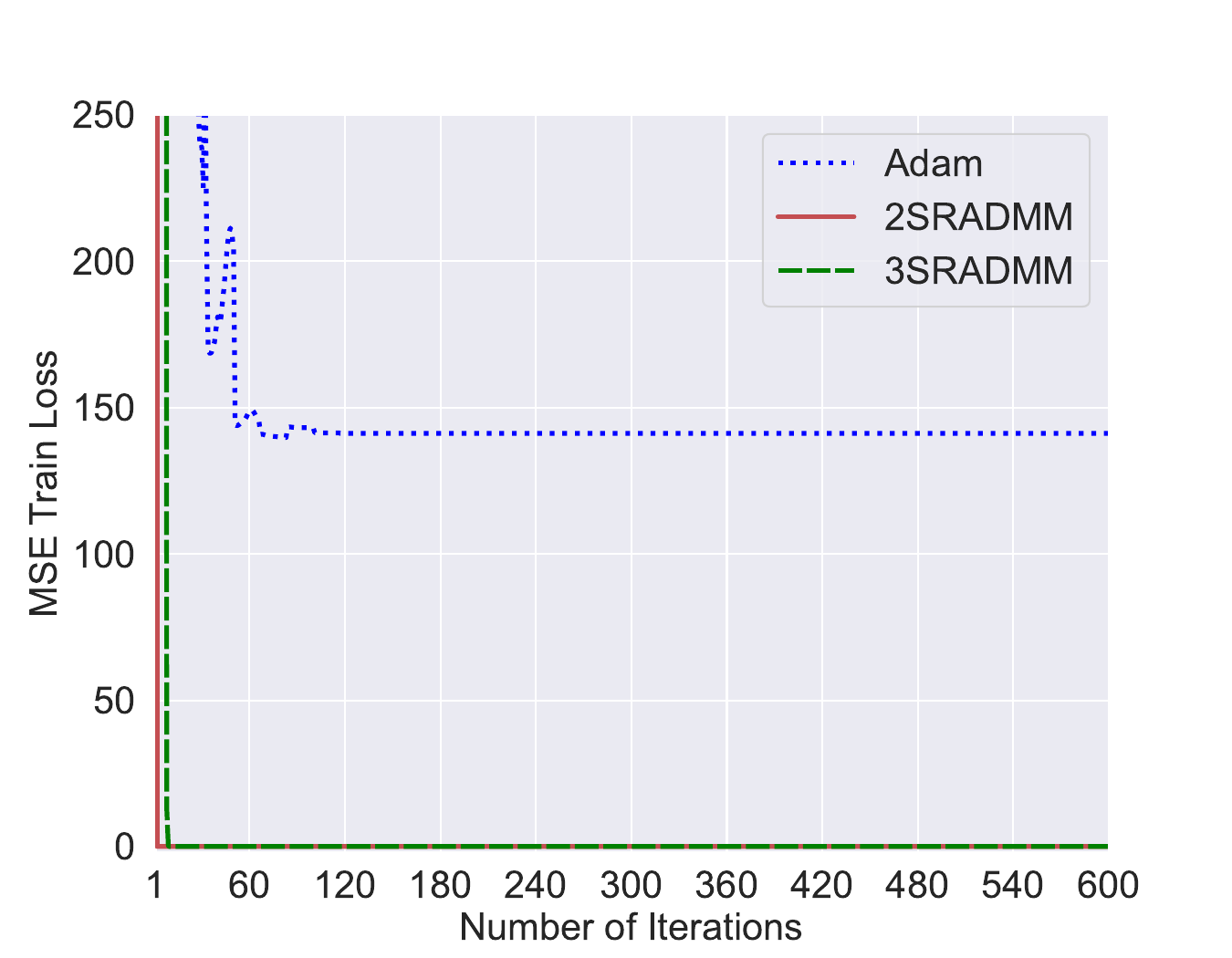}
\end{minipage}
\begin{minipage}[htb]{.48\linewidth}
\centering\label{fig:40_layers_sigmoid_WQ_test_MSE_loss}
\includegraphics[width=1\textwidth]{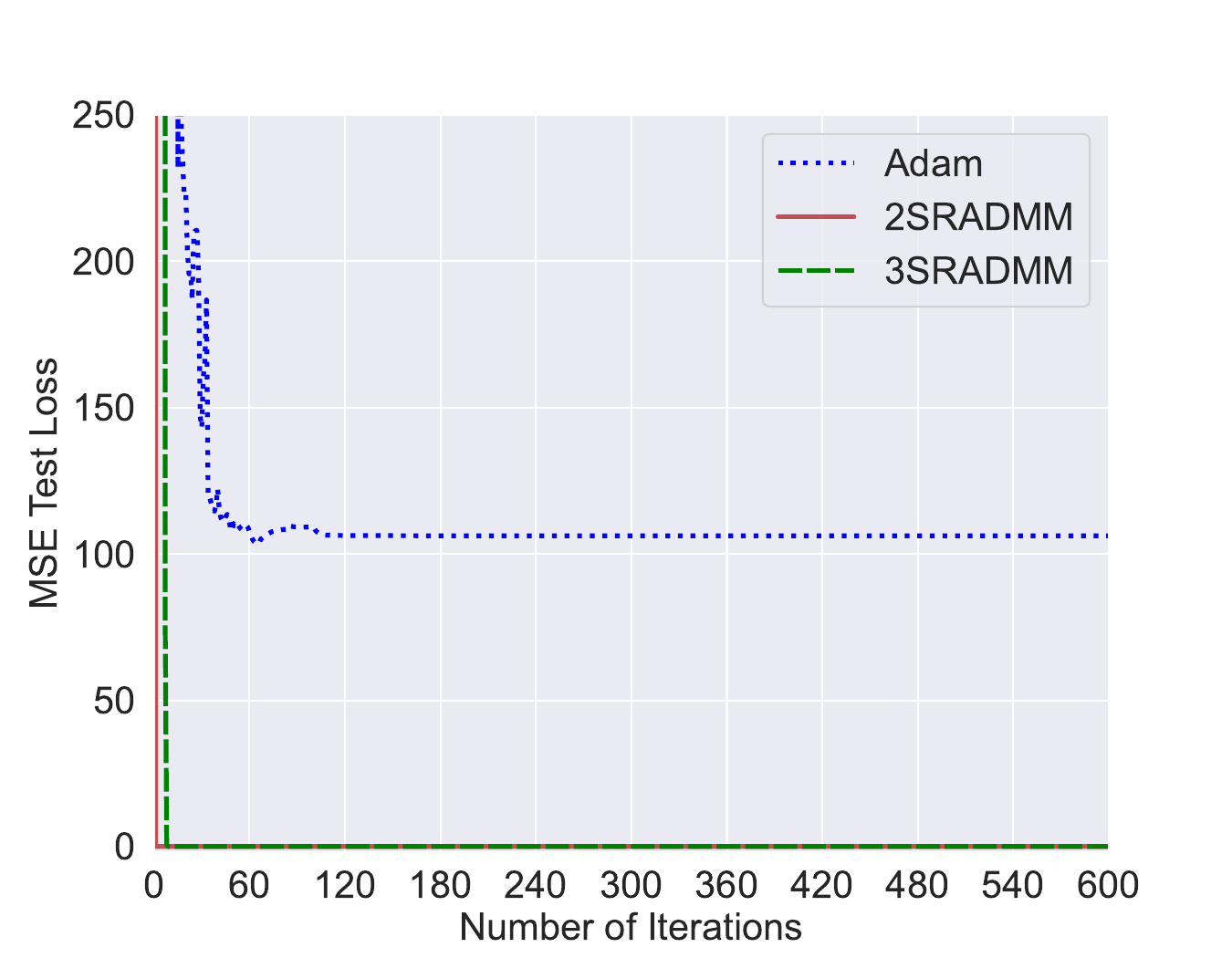}
\end{minipage}\\
\begin{minipage}[htb]{.48\linewidth}
\centering\label{fig:40_layers_sigmoid_WQ_test_MSE_loss}
\includegraphics[width=1\textwidth]{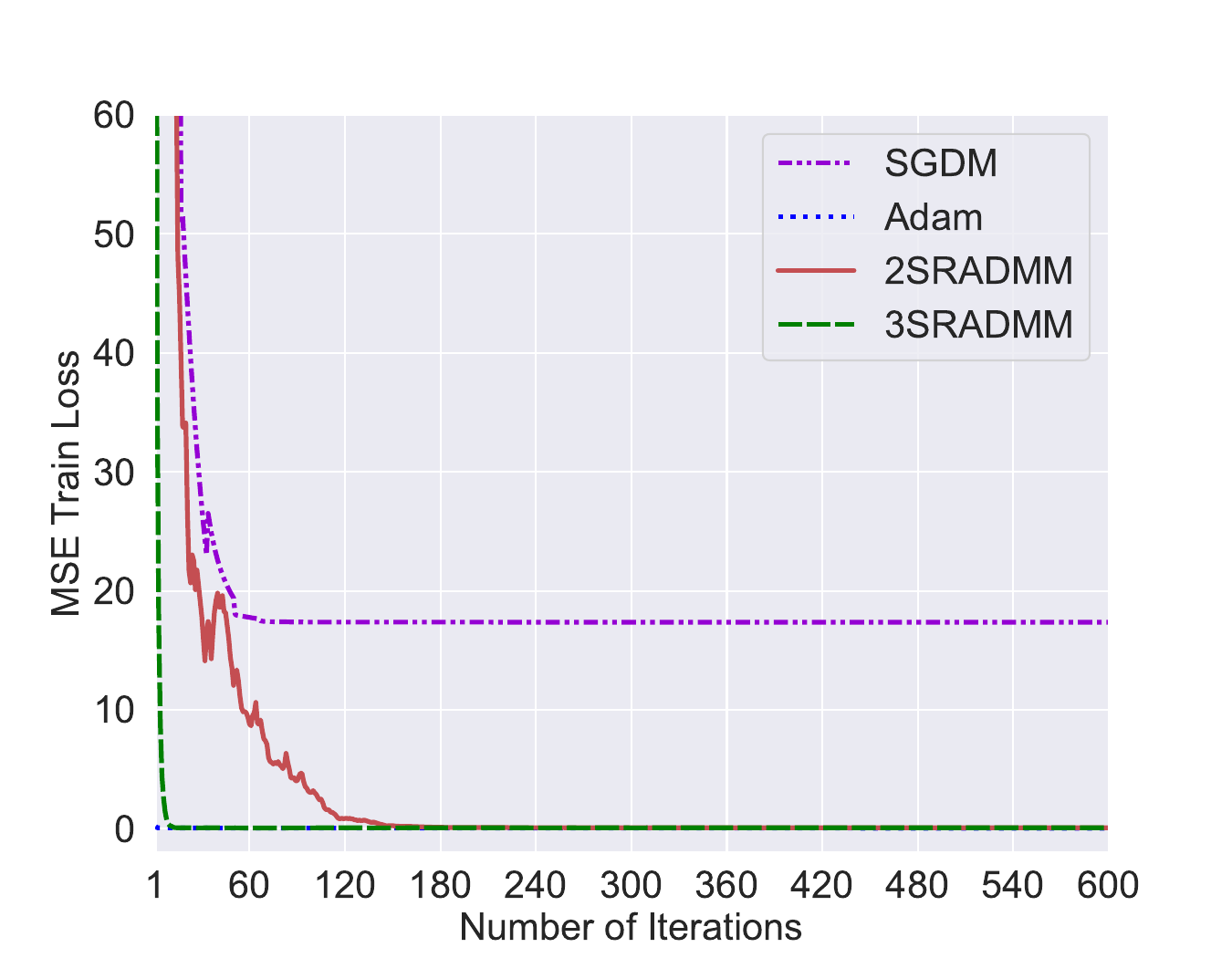}
\end{minipage}
\begin{minipage}[htb]{.48\linewidth}
\centering\label{fig:40_layers_sigmoid_WQ_test_MSE_loss}
\includegraphics[width=1\textwidth]{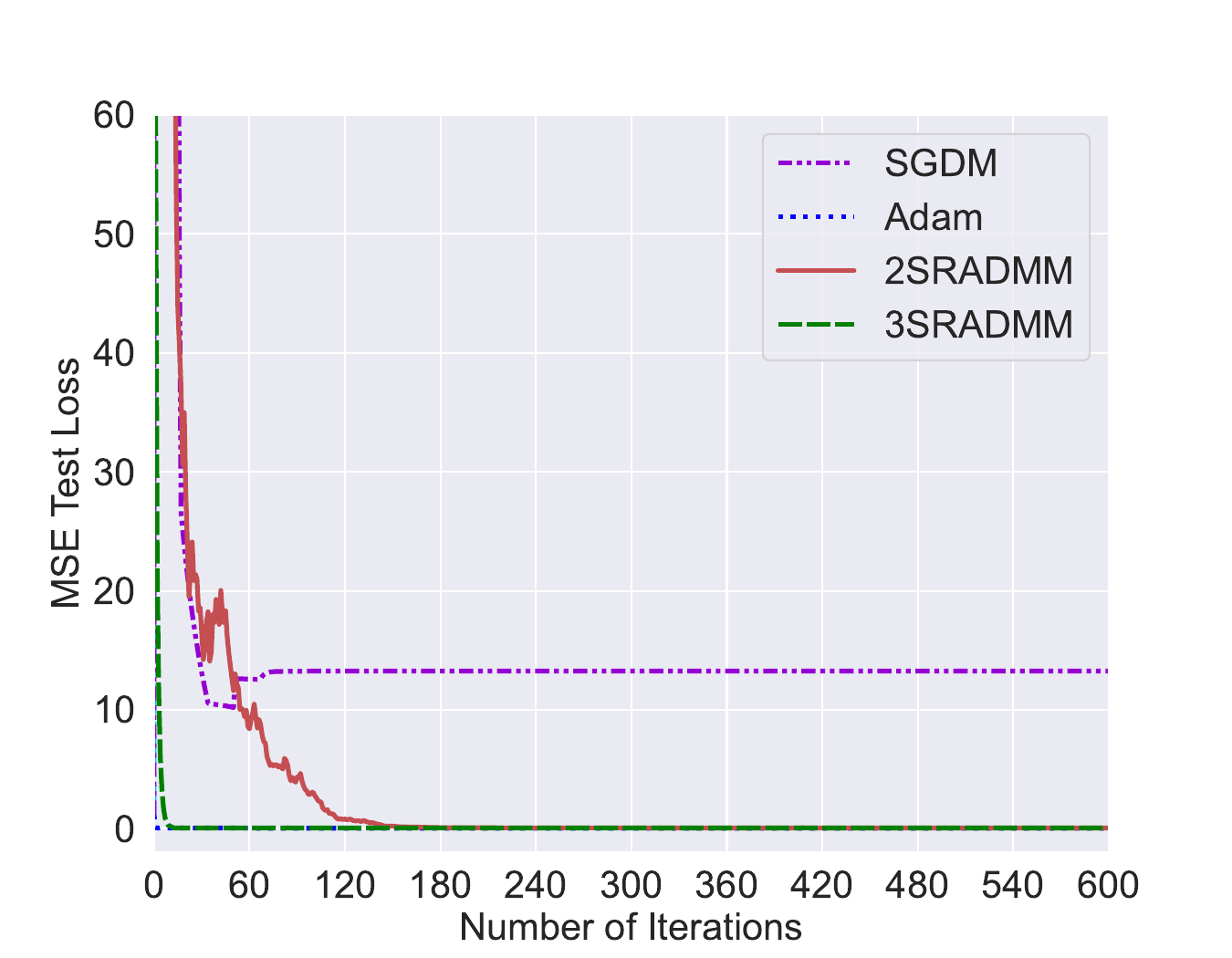}
\end{minipage}
\caption{MSE train (left), test (right) losses for the 40-layer ReLU (up), sigmoid (down) residual network on Wine Quality dataset.}
\end{figure}
\subsubsection{Superior Performance}
In this section, we compare the performance of our RADMM algorithms with SGD, SGDM, and Adam for training both relatively shallow and deep networks using ReLU or sigmoid activations, respectively. After 600 iterations, \Cref{table:better_performance} presents the MSE test loss for residual networks of various depths. The two smallest losses are highlighted in bold, and a hyphen (``-'') denotes an MSE test loss exceeding 500. For the ReLU network training task, our RADMMs significantly outperform the BP-based algorithms on deep networks (i.e., 20-, 30-, and 40-layer residual networks), with our three-splitting RADMM achieving the lowest MSE test loss on the 30- and 40-layer networks. For networks employing sigmoid activation, the three-splitting RADMM outperforms the two-splitting RADMM, and its performance is only marginally inferior to the best result obtained by Adam. In summary, benefiting from the ability to circumvent the exploding gradient issue, both the two- and three-splitting RADMM algorithms perform well on deep networks, with the three-splitting variant generally yielding lower MSE test losses than the two-splitting variant in most scenarios.
\vspace{-5mm}
{\footnotesize\begin{table}[h]
\centering
\caption{Mean Squared Error Test Losses}
\label{table:better_performance}
\tabcolsep = 1.2mm
\begin{tabular}{c|ccccc}
\hline
\diagbox{$N$}{Loss}{Algo} & SGD & SGDM & Adam & 2SRADMM & 3SRADMM\\
\hline
\multicolumn{6}{c}{~~~~~~~~~~~~~~~~ReLU}\\
\hline
3 & 0.2951 & 0.2951 & \textbf{0.0656} &  \textbf{0.0668} & 0.0941\\
10 & 301.7660 & 301.7660 & \textbf{0.0652} & 0.0657 & \textbf{0.0640}\\
20 & - & - & 0.1357 & \textbf{0.0705} & \textbf{0.0729}\\
30 & - &  - & 1.4892 & \textbf{0.0864} & \textbf{0.0708}\\
40 & - & - & 106.2628 & \textbf{0.0844} & \textbf{0.0725}\\
\hline
\multicolumn{6}{c}{~~~~~~~~~~~~~~~~Sigmoid}\\
\hline
3 & 0.0629 & 0.0586 & \textbf{0.0579} & 0.0593 & \textbf{0.0579}\\
10 & 0.0598 & \textbf{0.0596} & \textbf{0.0569} &  0.0606 & 0.0626 \\
20 & 2.2278 & 6.1277 & \textbf{0.0556} & 0.0848 & \textbf{0.0595}\\
30 & - & 0.0697  & \textbf{0.0574} &  0.1114 & \textbf{0.0647}\\
40 & - &  13.2671 &  \textbf{0.0577} & 0.0725 & \textbf{0.0688}\\
\hline
\end{tabular}
\end{table}}

\subsubsection{Enhanced Speed}
After 5 runs each with 600 iterations, the means and standard deviations of runtime of RADMMs, SGD, SGDM and Adam for 3-layer ReLU and sigmoid residual networks training on
Wine Quality dataset are shown in Figure 6, in which the standard deviation is reflected by the length of each error bar above and below the mean
value. Our two-splitting RADMM has the highest speed in the 5 algorithms on both ReLU and sigmoid networks. Besides, the runtimes of three-splitting RADMM on ReLU and sigmoid networks are acceptable and lower than those of Adam.
\vspace{-3mm}
\begin{figure}[htb]
\begin{minipage}[htb]{.49\linewidth}
\centering
\includegraphics[width=1\textwidth]{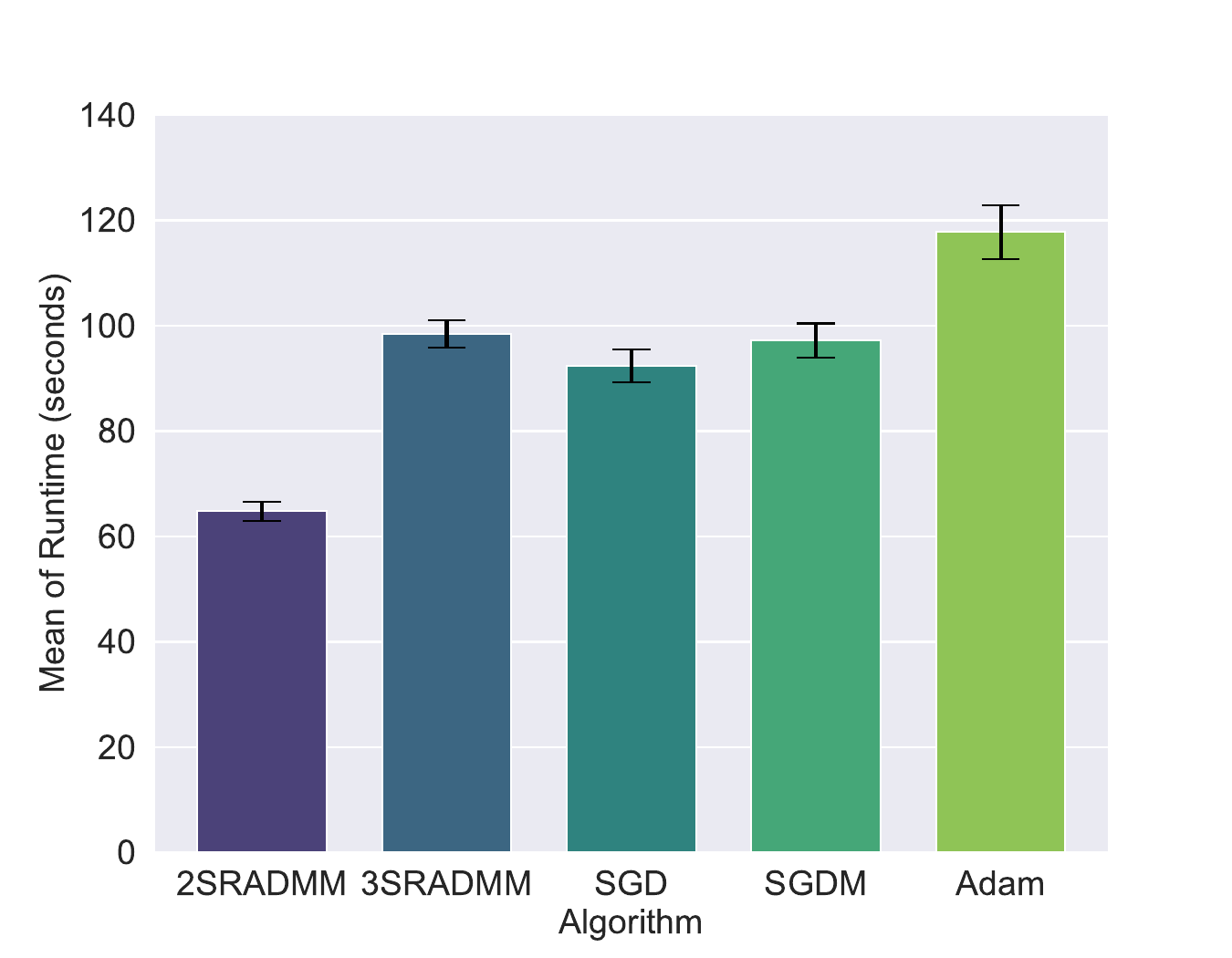}
\end{minipage}
\begin{minipage}[htb]{.49\linewidth}
\centering
\includegraphics[width=1\textwidth]{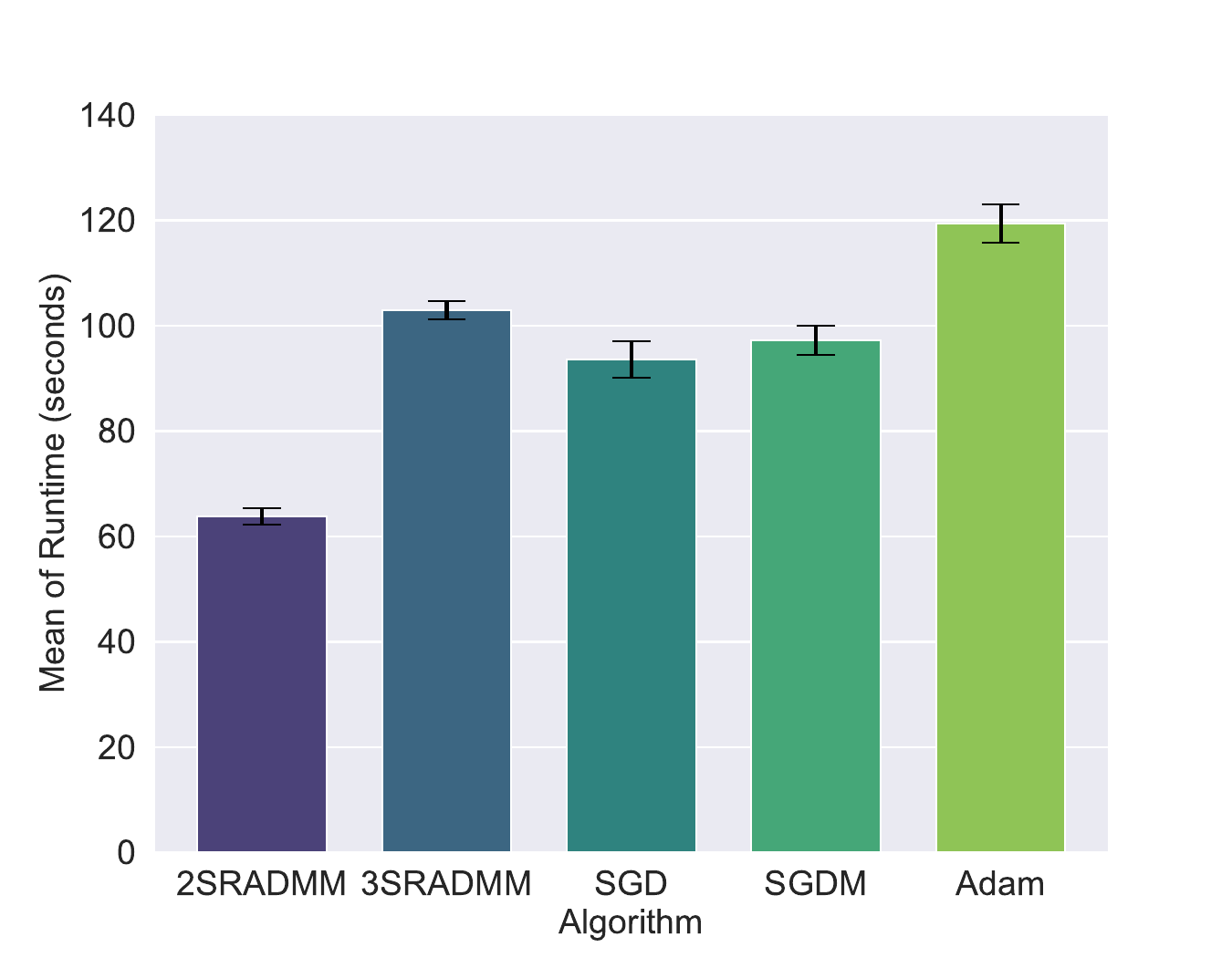}
\end{minipage}
\caption{Runtime for the 3-layer ReLU (left), sigmoid (right) residual
network on Wine Quality dataset.}
\end{figure}
\vspace{-2mm}
\subsection{Parallel Implementation}\label{subsec:parallel-implementation}
To better illustrate the time complexity benefits of parallelization, we design and implement a control protocol for the parallel proximal linearized RADMM algorithm (see Algorithm \ref{2sADMMp_prox}) using Python's multiprocessing and interprocess communication (IPC). In our implementation, the main procedure and the coordinator reside in a single process, while each network layer is assigned to its own subprocess. Communication among these processes is facilitated by message queues to transfer data and control signals. \Cref{fig:pipeline} presents a UML sequence diagram for a 3-layer network ($N=3$) as an example of this setup. Notably, our protocol can be easily extended to larger-scale tasks, such as multi-node clusters, by replacing the IPC mechanism with alternatives such as RPC.
\vspace{-3.5mm}
\begin{figure}[h]
  \centering
  \includegraphics[width=1\linewidth]{split2-parallel.pdf}
\caption{UML sequence diagram for PRU parallel training, taking a 3-layer residual network as an example. $X$, $Y$: the data and label in the current batch; message \texttt{STEP}: begin training on the next batch; message \texttt{TICK}: sent from layer (process) $i$ to layer (process) $i+1$ after updating $V_i$; message \texttt{TOCK}: sent from layer (process) $i+1$ to layer (process) $i$ after updating $W_i$; message \texttt{SYNC}: after completing \texttt{n\_updates} updates on a batch, all processes synchronize with the main procedure and return results; message \texttt{QUIT}: training is complete, notify all processes to terminate.}
		\label{fig:pipeline}
\end{figure}
The oscillating function below is fitted using a 5-layer network with sigmoid activation.

\vspace{-3mm}
{\small
\begin{align*}
\begin{split}\label{oscilation function}
f(\x)=
\begin{cases}
~ x_1x_2^2...x_{d-1}x_d^2, &~~ \x\in(-\infty, -1]^d;\\
~ x_1^2x_2^2...x_{d-1}^2x_d^2, &~~ \x\in(-\infty, -1]^d-(-\infty, -1]^d;\\
~ x_1^2x_2...x_{d-1}^2x_d, &~~ \x\in\{x_1>1, \ldots, \text{or}~x_d>1\}.
\end{cases}
\end{split}
\end{align*}}

\noindent As noted in \Cref{subsec:lower-time-complexity}, parallelization reduces the computation cost, albeit at the expense of additional communication overhead. In our experiments, the network width $d$
predominantly governs the overall computation cost, since $T_{\scriptscriptstyle \rm mul}$ scales super-linearly with $d$
\cite{Williams2024}, while the communication cost remains nearly constant in a shared-memory configuration. Each test case was executed 5 times, and \ref{table:parallel implementation} presents the mean runtime (in seconds) along with the corresponding standard deviations.
{\footnotesize\begin{table}[h]
\centering
\caption{Parallel vs. serial RADMMs in runtime.}
\label{table:parallel implementation}
\begin{tabular}{ccc}
\hline
$d$& Serial RADMM & Parallel RADMM\\
\hline
3000 & ~~~~\textbf{253.8010}\scriptsize{~$\pm$1.0088} & ~~~~313.3090\scriptsize{~$\pm$1.4582}\\
4000 & ~~~~\textbf{543.0684}\scriptsize{~$\pm$2.3043} & ~~~~603.5077\scriptsize{~$\pm$1.5736}\\
5000 & ~~~~1132.5736\scriptsize{~$\pm$13.3211} & ~~~~\textbf{1110.6362}\scriptsize{~$\pm$10.9119}\\
5500 & ~~~~1650.6885\scriptsize{~$\pm$24.2471} & ~~~~\textbf{1551.2158}\scriptsize{~$\pm$107.8684}\\
6000 & ~~~~2314.2873\scriptsize{~$\pm$51.1496} & ~~~~\textbf{1968.3845}\scriptsize{~$\pm$51.0245}\\
6500 & ~~~~2711.8983\scriptsize{~$\pm$19.6904} & ~~~~\textbf{2483.0785}\scriptsize{~$\pm$59.7868}\\
\hline
\end{tabular}
\end{table}}
The results indicate that when the network width
$d \ge 5000$, the runtime of the parallel implementation is lower than that of the serial implementation, as the computation cost becomes the dominant factor rather than communication overhead. Moreover, the performance advantage of the parallel implementation increases with the network dimension. These findings suggest that our parallel algorithm holds substantial potential for large-scale training problems.

\section{CONCLUSION}\label{sec:concluding_remarks}
In this paper, we propose both serial and parallel proximal (linearized) ADMM-based training algorithms for residual networks, leveraging a parallel regional update mechanism. We establish theoretical guarantees for convergence and demonstrate that our algorithms achieve an R-linear (or sublinear) convergence rate. Moreover, our analysis reveals that the parallel (distributed) RADMM algorithms offer advantages in terms of reduced time complexity and lower memory consumption. Additionally, we design and implement a control protocol for the parallel RADMMs using Python's multiprocessing and IPC. Experimental results demonstrate the fast and stable convergence, superior performance, and enhanced computational speed of our RADMM algorithms in training residual networks, further underscoring the benefits of the parallel implementation.

Potential extensions of this work include incorporating randomness into the data to design stochastic ADMMs, as well as developing and analyzing ADMM training algorithms for additional networks, such as CNNs and Transformers, to address image and natural language processing tasks. Finally, although our proof-of-concept implementation confirms the feasibility of our approach, the relatively high communication overhead in the parallel implementation partially offsets the benefits of parallelization, making the optimization of communication strategies a direction for future research.

\section{APPENDIX}\label{sec:appendix}
\subsection{Values of Parameters in Assumption \ref{assum:3s_prox_assumption2}}\label{subsec:values}
\vspace{-4mm}
{\footnotesize\begin{align*}
\bullet~&\underline{\beta}_ {i}=\max\Big\{32(1+\sqrt{2})\mu(\psi_{0}\psi_{2}+\psi_{1}^{2}+\overline{\mathcal{V}}_{i}\psi_{2}+\overline{\mathcal{V}}_{i-1}\psi_{2}),
16\mu \psi_{1}^{2}\Big\},\\
\bullet~&\underline{\beta}_{N}=1,
\end{align*}}
\vspace{-4mm}

\noindent for some $\overline{\mathcal{V}}_{0}\in(\Vert \X\Vert, +\infty)$ and $\{\overline{\mathcal{V}}_{i}\}_{i=1}^{N-1}\subseteq(0, +\infty)$,
\vspace{-1.5mm}
{\footnotesize\begin{align*}
\footnotesize
\bullet~ \underline{\omega}_{i}:=\frac{1}{16}\big(\frac{\beta_{i}}{4}-8\mu(\psi_{0}\psi_{2}+\psi_{1}^{2}+\overline{\mathcal{V}}_{i}\psi_{2}+\overline{\mathcal{V}}_{i-1}\psi_{2})-\sqrt{\Delta_{i}}\big)+\hat{\epsilon}_i,
\end{align*}}
\vspace{-2mm}
\noindent
where
{\footnotesize\begin{align*}
\Delta_ {i}\coloneqq&\big(\frac{\beta_{i}}{4}-8\mu(\psi_{0}\psi_{2}+\psi_{1}^{2}+\overline{\mathcal{V}}_{i}\psi_{2}+\overline{\mathcal{V}}_{i-1}\psi_{2})\big)^{2}\\
&-128\mu^{2}(\psi_{0}\psi_{2}+\psi_{1}^{2}+\overline{\mathcal{V}}_{i}\psi_{2}+\overline{\mathcal{V}}_{i-1}\psi_{2})^{2},
\end{align*}}

\noindent and $\hat{\epsilon}_i\in(0, \frac{\sqrt{\Delta_i}}{32})$, $i\in[N-1]$,
\vspace{-1.5mm}
{\footnotesize\begin{align*}
\bullet~\overline{\omega}_i&\coloneqq\min\Bigg\{\frac{\frac{\beta_{i}}{4}-8\mu(\psi_{0}\psi_{2}+\psi_{1}^{2}+\overline{\mathcal{V}}_{i}\psi_{2}+\overline{\mathcal{V}}_{i-1}\psi_{2})+\sqrt{\Delta_{i}}}{16}-\hat{\epsilon}_i,\\
&~~~~~~~~~~~~\sqrt{(\underline{\omega}_{i})^{2}+\frac{\beta_i\underline{\omega}_{i}}{16}-\frac{\beta_i\epsilon_i}{4}}\Bigg\},~~~~~~~~~~~~~~~~~~
\end{align*}}
\vspace{-3.5mm}

\noindent in which $\epsilon_i\in(0, \frac{\underline{\omega}_i}{4}), i\in[N-1]$ with the $\{\overline{\mathcal{V}}_i\}_{i=0}^{N-1}$.
\subsection{Parallel Three-Splitting RADMM}
Pseudocode of our parallel three-splitting RADMM training algorithm is shown in Algorithm \ref{alg:3sADMM_p}.
{\footnotesize\begin{algorithm}[h]\label{alg:3sADMM_p}
\footnotesize
\caption{Parallel Three-Splitting RADMMs}
\SetKwInput{Initialization}{Initialization}
\SetKw{Return}{return}
\SetKwFor{ForParallel}{parallel\_for}{do}{end}
\SetKwFunction{Range}{range}
\SetKwInput{Initialization}{Initialization}
\SetKwInput{Input}{Input}
\SetKwInput{Output}{Output}

\Input{$\X$, $\Y$, $K$, $\lambda$, $\mu$, $\{\beta_{i}\}_{i=1}^N$, $\{\omega_i^k\}_{i=1}^{N-1}$, and $\{\W_{i}^{0}\}_{i=1}^N$.}
$\V_{0}^{k}\gets \X$, $\U_{i}^{0}\gets \W_{i}^{0}\V_{i-1}^{0}$, $\V_{i}^{0}\gets \V_{i-1}^{0}+\sigma_i(\U_{i}^{0})$, $i\in[N-1]$, $\V_{N}^{0}\gets \W_{N}^{0}\V_{N-1}^{0}$ and $\Lam_{i}^{0}\gets \boldsymbol{0}$, $i\in[N]$.

\ForParallel{$i\in[N]$}{
    \For{$k\gets0$~{\rm\textbf{to}}~$K-1$}{
    \eIf{$i < N$}{
           Update $\W_{i}^{k+1}$ by iteration equation (\ref{eq:3s-Wi-update}).
        }{
        Update $\W_{N}^{k+1}$ by iteration equation (\ref{eq:3s-Wn-update}).
        }
        \lIf{$i > 1$}{Retrieve necessary block variables from processor $i-1$}
        \If{$i < N$}{
            Update $\U_{i}^{k+1}$ by solving (\ref{eq:3s-prox-Ui-update}) in proximal
ADMM (by iteration equation (\ref{eq:3s-proxgra-Ui-update}) in proximal linearized ADMM).
        }

        \lIf{$i < N$}{Retrieve necessary block variables from processor $i+1$}
        \uIf{$i<N-1$}{Update $\V_{i}^{k}$ by iteration equation (\ref{eq:3s-Vi-update}).
        }
        \uElseIf{$i=N-1$}{
        Update $\V_{N-1}^{k+1}$ by iteration equation (\ref{eq:3s-Vn-1-update}).
        }
        \Else{
        Update $\V_{N}^{k+1}$ by iteration equation (\ref{eq:3s-Vn-update}).
        }

        \eIf{$i < N$}{
            Update $\Lam_{i}^{k+1}$ by iteration equation (\ref{eq:3s-lambdai-update}).
        }{
            Update $\Lam_{i}^{k+1}$ by iteration equation (\ref{eq:3s-lambdan-update}).
        }
    }
}
Synchronize all processors.\\
\Output{$\{\W_{i}^K\}_{i=1}^{N}$}
\end{algorithm}}
\subsection{Experimental Settings}\label{subsec:experimental settings}
Experiments in \Cref{subsec:wine quality dataset} and \Cref{subsec:parallel-implementation} are conducted on a laptop equipped with an Intel Core i5-11320H @ 3.20GHz CPU (4 cores 8 threads), and a server equipped with two Intel Xeon Gold 5218 @ 2.30GHz CPUs (16 cores 32 threads per socket), respectively. Parameters are initialized with a Kaiming normal distribution \cite{He2015}. Training and testing data in \Cref{subsec:wine quality dataset} are normalized as $\frac{x - x_{\min}}{x_{\max} - x_{\min}}$.

Parameters used in \Cref{subsec:wine quality dataset} are listed as below, in which expansion factor \texttt{EF}$_{\tau}$, \texttt{EF}$_{\iota}$ are multiplied to $\tau, \iota$ after each epoch, respectively.

\noindent $\bullet$ Sigmoid. BP-based algorithms: learning rate $=0.01$, weight decay $=0.0001$, momentum $=0.9$. 2SRADMM: \texttt{EF}$_{\tau}=1.05$, \texttt{EF}$_{\iota}=1.05$, $\beta= 10000$, $\mu=0.1$, $\lambda=$ 0.05 for $N \le 10$, 5 for $10< N \le 30$, 50 otherwise, initial $\tau_i\equiv N$, $\iota_i\equiv N$. 3SRADMM: \texttt{EF}$_{\tau}=1.05$, $\beta_i \equiv 1000$, $\mu = 0.1$, $\lambda = 0.0001$, initial $\tau_i\equiv 10$.

\noindent $\bullet$ ReLU. BP-based algorithms: SGD learning rate$=10^{-10}/\delta d$, SGD weight decay $=10^{-10}\delta d$, where $\delta = 10^{\lfloor 0.2N\rfloor}$, SGD momentum $=0.9$, Adam learning rate $=0.01$, adam weight decay $=1$. 2SRADMM: \texttt{EF}$_{\tau}=1.05$, \texttt{EF}$_{\iota}=1.05$, $\beta= 10$, $\mu = 10^{-5} / dN$, $\lambda = 10^{-5}dN$, $\tau_i\equiv 10dN$, initial $\iota_i\equiv 10d$. 3SRADMM: \texttt{EF}$_{\tau}=1.05$, $\beta_i \equiv 100$, $\mu = 1$, $\lambda = 0.0001$, initial $\tau_i\equiv 10$.

In addition, 10 samples $\{(\x_i, f(\x_i))\}_{i=1}^{10}$ are produced uniformly in $[-2, 2)$, and we take $\beta=100$, $\mu=0.1$, $\tau_i^k\equiv 1$, $\iota_i^k\equiv1$, $\lambda=0.005$ for $d=3000, 4000$, and $0.001$ for $d=5000, 5500, 6000, 6500$ in the two-splitting proximal linearized RADMM in \Cref{subsec:parallel-implementation}.

\vspace{-2.5mm}
\subsection{Proof of Proposition \ref{prop:2s_prox_theorem1}}\label{app:1}
We first define $\Theta_{\X_i^{k+1}}$ as {\small$(\X_1^{k+1}, \ldots, \X_{i-1}^{k+1}, \X_{i+1}^k, \ldots, \X_{n}^k)$}.

\begin{IEEEproof}{(Proof of Proposition \ref{prop:2s_prox_theorem1})}
For the update of $\W_N^k$,
{\small\begin{align*}
&\mathcal{L}_{\beta}^{2s}(\W_{N}^{k+1};\Theta_{\W_N^{k+1}}) - \mathcal{L}_{\beta}^{2s}(\W_{N}^{k};\Theta_{\W_N^{k}})\\
=&-\frac{\lambda}{2}\Vert \W_{N}^{k+1}-\W_{N}^{k}\Vert^{2}-\frac{\beta}{2}\Vert (\W_{N}^{k+1}-\W_{N}^{k})\V_{N-1}^{k}\Vert^{2}
\end{align*}}
\vspace{-3.5mm}

\noindent by the first-order optimality condition of (\ref{eq:2s-Wn-update}).
For the update of each $\W_i^k$, $i\in[N-1]$, we have
\vspace{-2mm}
{\small\begin{align*}
\mathcal{L}_{\beta}^{2s}(\W_{i}^{k+1};\Theta_{\W_i^{k+1}})-\mathcal{L}_{\beta}^{2s}(\W_{i}^{k};\Theta_{\W_i^{k}})\leq-\frac{\omega_{i}^{k}}{2}\Vert \W_{i}^{k+1}-\W_{i}^{k}\Vert^{2}.
\end{align*}}
\vspace{-4mm}

\noindent Similarly, we can estimate the change of $\mathcal{L}_{\beta}^{2s}$ related to $\V_i^k$, $i\in [N-1]$.
For the update of $\V_N^k$, we have
{\small\begin{align*}
\mathcal{L}_{\beta}^{2s}(\V_{N}^{k+1};\Theta_{\V_N^{k+1}})-\mathcal{L}_{\beta}^{2s}(\V_{N}^{k};\Theta_{\V_N^{k}})\leq-\frac{1+\beta}{2}\Vert \V_{N}^{k+1}-\V_{N}^{k}\Vert^{2}
\end{align*}}
\vspace{-4mm}

\noindent based on the strong convexity of the objective function of (\ref{eq:2s-Vn-update}).

\noindent For the update of $\Lam^{k}$, we have
\vspace{-2mm}
{\small\begin{align*}
\mathcal{L}_{\beta}^{2s}(\Lam^{k+1};\Theta_{\Lam^k+1})
=\mathcal{L}_{\beta}^{2s}(\Lam^{k};\Theta_{\Lam^{k}})+\frac{1}{\beta}\Vert\Lam^{k+1}-\Lam^{k}\Vert^{2}.
\end{align*}}
\vspace{-5mm}

\noindent Adding the above difference of adjacent iterations in the update of each block variable, we have
\vspace{-0.5mm}
{\small
\begin{align*}
&\mathcal{L}_{\beta}^{2s}(\Psi_{2s}^{k+1})-\mathcal{L}_{\beta}^{2s}(\Psi_{2s}^{k})&~\\
\leq& -\frac{\lambda}{2}\Vert \W_{N}^{k+1}-\W_{N}^{k}\Vert^{2}-\sum_{i=1}^{N-1}\frac{\underline{\omega}_{i}}{2}\Vert \W_{i}^{k+1}-\W_{i}^{k}\Vert^{2}\\
&-\sum_{i=1}^{N-2}\frac{\underline{\nu}_{i}}{2}\Vert \V_{i}^{k+1}-\V_{i}^{k}\Vert^{2}-\frac{\mu}{2}\Vert \V_{N-1}^{k+1}-\V_{N-1}^{k}\Vert^{2}\\
&-\big(\frac{1+\beta}{4}-\frac{1}{2\beta}\big)\big(\Vert \V_{N}^{k+1}-\V_{N}^{k}\Vert^{2}+\Vert \Lam^{k+1}-\Lam^{k}\Vert^{2}\big)\\
\leq&  -c_1\Vert \Psi_{2s}^{k+1}-\Psi_{2s}^{k}\Vert^{2},
\end{align*}}

\vspace{-2mm}
\noindent where
$c_1=\min\{\frac{\lambda}{2}, \{\frac{\omega_{i}^{\min}}{2}\}, \{\frac{\nu_{i}^{\min}}{2}\}, \mu, \frac{1+\beta}{4}-\frac{1}{2\beta}\}>0$.
\end{IEEEproof}
\vspace{-2mm}
\subsection{Proof of Proposition \ref{prop:2s_prox_theorem2}}\label{app:2}
First, upper boundness of
$\{\Vert \W_{i}^{k}\Vert\}_{i=1}^N$, $\{\Vert \V_{i}^{k}\Vert\}_{i=1}^N$, $\{\Vert\Lam^{k}\Vert\}$ under Assumptions \ref{assum:assumption2s2} is guaranteed in \Cref{2s_prox_lemma6}.
\begin{lemma}\label{2s_prox_lemma6}
There exist positive constants $\{\overline{\mathcal{W}}_{i}\}_{i=1}^N$, $\{\overline{\mathcal{V}}_{i}\}_{i=1}^N$, $\overline{\mathcal{\lambda}}$ such that $\Vert \W_{i}^{k}\Vert\leq\overline{\mathcal{W}}_{i}$, $\Vert \V_{i}^{k}\Vert\leq\overline{\mathcal{V}}_{i}$, $i\in[N]$, $\Vert\Lam^{k}\Vert\leq\overline{\mathcal{\lambda}}$.
\end{lemma}
\begin{IEEEproof}
By (\ref{eq:2s-lambda-update}), we have
\vspace{-2mm}
{\small\begin{align*}
&(\frac{1}{2}-\frac{1}{2\beta})\Vert \V_{N}^{k}-\Y\Vert^{2}+\frac{\mu}{2}\sum_{i=1}^{N-1}\Vert \V_{i-1}^{k}+\sigma_i\big(\W_{i}^{k}\V_{i-1}^{k}\big)-\V_{i}^{k}\Vert^{2}\\[-0.1bp]
&+\frac{\lambda}{2}\sum_{i=1}^{N}\Vert \W_{i}^{k}\Vert^{2}+\frac{\beta}{2}\Vert \W_{N}^{k}\V_{N-1}^{k}-\V_{N}^{k}+\frac{1}{\beta}\Lam^{k}\Vert^{2}<\infty
\end{align*}}

\vspace{-4mm}
\noindent by \Cref{prop:2s_prox_theorem1}.
If $\Vert \W_{i}^{k}\Vert\to\infty$, then $\mathcal{L}_{\beta}^{2s}(\Psi_{2s}^ {k})\to\infty$, a contradiction. Thus there exist $\overline{\mathcal{W}}_{i}>0$ such that $\Vert \W_{i}^{k}\Vert\leq\overline{\mathcal{W}}_{i}$, $i\in [N]$.
If $\Vert \V_{N}^{k}-\Y\Vert_{F}\to\infty$, then $\mathcal{L}_{\beta}^{2s}(\Psi_{2s}^ {k})\to\infty$, a contradiction. Thus there exist $\overline{\mathcal{V}}_{N}$ and $ \overline{\mathcal{\lambda}}>0$ such that $\Vert \V_{N}^{k}\Vert_{F}\leq\overline{\mathcal{V}}_{N}$, $\Vert\Lam^{k}\Vert_{F}\leq\overline{\mathcal{\lambda}}$. Similarly, the sequences $\{\Vert \V_{i}^{k}-\V_{i-1}^{k}\Vert_{F}\}, i\in[N-1]$ are also upper bounded. By $\V_{0}^{k}\equiv \X$, there exist $\overline{\mathcal{V}}_{i}>0$ such that $\Vert \V_{i}^{k}\Vert_{F}\leq\overline{\mathcal{V}_{i}}, i\in[N-1]$.
\end{IEEEproof}
\vspace{2mm}
\begin{IEEEproof}{(Proof of Proposition \ref{prop:2s_prox_theorem2})}
By the first-order optimality condition of (\ref{eq:2s-Wn-update}), we know that
\vspace{-1mm}
{\footnotesize\begin{align*}
&\frac{\partial \mathcal{L}_{\beta}^{2s}}{\partial \W_{N}}(\Psi_{2s}^{k+1})
=\beta\{\W_{N}^{k+1}\V_{N-1}^{k+1}\V_{N-1}^{k+1~\mathrm{T}}-\W_{N}^{k+1}\V_{N-1}^{k}\V_{N-1}^{k~\mathrm{T}}\\
&+\V_{N}^{k}\V_{N-1}^{k~\mathrm{T}}-\V_{N}^{k+1}\V_{N-1}^{k+1~\mathrm{T}}\}+\Lam^{k+1}\V_{N-1}^{k+1~\mathrm{T}}-\Lam^{k}\V_{N-1}^{k~\mathrm{T}}.
\end{align*}}

\vspace{-2mm}
\noindent Thus we have
\vspace{-3mm}
{\small\begin{align*}
&\Big\Vert\frac{\partial \mathcal{L}_{\beta}^{2s}}{\partial \W_{N}}(\Psi_{2s}^{k+1})\Big\Vert
\leq\overline{\mathcal{V}}_{N-1}\Vert\Lam^{k+1}-\Lam^{k}\Vert+\beta\overline{\mathcal{V}}_{N-1}\Vert \V_{N}^{k+1}-\V_{N}^{k}\Vert\\
&+(2\beta\overline{\mathcal{W}}_{N}\overline{\mathcal{V}}_{N-1}+\beta\overline{\mathcal{V}}_{N}+\overline{\mathcal{\lambda}})\Vert \V_{N-1}^{k+1}-\V_{N-1}^{k}\Vert.
\end{align*}}

\vspace{-1mm}
\noindent Similarly, we can estimate the norms of partial derivative of $\mathcal{L}_\beta^{2s}$ with respect to $\W_i$, $\V_i$, and $\Lam$, respectively. Then
\vspace{-1mm}
{\small\begin{align*}
&\Vert\nabla \mathcal{L}_{\beta}^{2s}(\Psi_{2s}^{k+1})\Vert\\
\leq&\sum_{i=1}^{N}\Big\Vert\frac{\partial \mathcal{L}_{\beta}^{2s}}{\partial \W_{i}}(\Psi_{2s}^{k+1})\Big\Vert+\sum_{i=1}^{N}\Big\Vert\frac{\partial \mathcal{L}_{\beta}^{2s}}{\partial \V_{i}}(\Psi_{2s}^{k+1})\Big\Vert+\Big\Vert\frac{\partial \mathcal{L}_{\beta}^{2s}}{\partial \Lam}(\Psi_{2s}^{k+1})\Big\Vert\\
\leq& \tilde c(\sum_{i=1}^{N}\Vert \W_{i}^{k+1}-\W_{i}^{k}\Vert+\sum_{i=1}^{N}\Vert \V_{i}^{k+1}-\V_{i}^{k}\Vert+\Vert\Lam^{k+1}-\Lam^{k}\Vert)\\
\leq &c_{2}\Vert \Psi_{2s}^{k+1}-\Psi_{2s}^{k}\Vert
\end{align*}}

\vspace{-1mm}
\noindent for certain $\tilde c>0$, and $c_{2}\coloneqq\sqrt{2N+1}\tilde c>0$.
\end{IEEEproof}
\vspace{-5mm}
\subsection{Proofs of Theorems \ref{thm:thm_2s_prox_main_result1} and \ref{thm:3s_prox_theorem6}}\label{proof of theorems 1 and 2}
\begin{IEEEproof}{(Proof of Theorems \ref{thm:thm_2s_prox_main_result1} and \ref{thm:3s_prox_theorem6})}
Based on Propositions \ref{prop:2s_prox_theorem1} and \ref{prop:2s_prox_theorem2}, following from Theorem 2.9 of \cite{Attouch2013}, Theorem 2 of \cite{Attouch2009} (Lemma 5 of \cite{Baque2015}), and Theorem 1 of \cite{Xu2022}, we obtain the convergence results.
\end{IEEEproof}
\vspace{-4mm}
\subsection{Proof of Lemma \ref{lem:1}}\label{proof of lemma 1}
Under the assumptions mentioned in \Cref{subsubsec:convergence results}, we first prove the sufficient descent of $\mathcal{L}_R^{3s}$ and estimate the norm of gradient $\Vert \nabla\mathcal{L}_{R}^{3s}(\widetilde\Psi_{3s}^{k})\Vert$ as in \Cref{lemma:3s_inter_result} and \Cref{3s_prox_theorem1}.
\begin{lemma}\label{lemma:3s_inter_result}
The serial and parallel three-splitting RADMMs satisfy \Cref{prop:2s_prox_theorem1} with respect to $\mathcal{L}_R^{3s}$ and $\widetilde\Psi_{3s}$.
\end{lemma}
\begin{IEEEproof}
Using the method similar to the proof of Proposition \ref{prop:2s_prox_theorem1}, we have
{\footnotesize\begin{align*}
&\mathcal{L}_{\beta}^{3s}(\Psi_{3s}^{k+1})+\sum_{i=1}^{N-1}(\frac{4\big(\underline{\omega}_i)^2}{\beta_i}+\frac{\underline{\omega}_i}{4})\Vert \U_{i}^{k+1}-\U_{i}^{k}\Vert^{2}\\
&+\sum_{i=1}^{N-1}(\tilde{\eta}_{i}+\frac{\mu}{4})\Vert \V_{i}^{k+1}-\V_{i}^{k}\Vert^{2}\\
\leq& \mathcal{L}_{\beta}^{3s}(\Psi_{3s}^{k})+\sum_{i=1}^{N-1}(\frac{4(\underline{\omega}_i)^2}{\beta_i}+\frac{\underline{\omega}_i}{4})\Vert \U_{i}^{k}-\U_{i}^{k-1}\Vert^{2}\\
&+\sum_{i=1}^{N-1}(\tilde{\eta}_{i}+\frac{\mu}{4})\Vert \V_{i}^{k}-\V_{i}^{k-1}\Vert^{2}-\sum_{i=1}^{N-1}(\bar{\eta}_{i}-\tilde{\eta}_{i}-\frac{\mu}{4})\Vert \V_{i}^{k+1}-\V_{i}^{k}\Vert^{2}\\
&-\sum_{i=1}^{N-1}(\hat{\theta}_{i}^{k+1}-\tilde{\theta}_{i}^{k}-\frac{\omega_{i}^{k-1}}{4})\Vert \U_{i}^{k+1}-\U_{i}^{k}\Vert^{2}-\sum_{i=1}^{N-1}\frac{\mu}{4}\Vert \V_{i}^{k}-\V_{i}^{k-1}\Vert^{2}\\
&-\sum_{i=1}^{N-1}(\frac{4(\underline{\omega}_i)^2}{\beta_i}+\frac{\underline{\omega}_i}{4}-\tilde{\theta}_{i}^{k}) \Vert \U_{i}^{k}-\U_{i}^{k-1}\Vert^{2}\\
&-\frac{\lambda}{2}\sum_{i=1}^{N}\Vert \W_{i}^{k+1}-\W_{i}^{k}\Vert^{2}-(\frac{1+\beta_{N}}{2}-\frac{1}{\beta_{N}})\Vert \V_{N}^{k+1}-\V_{N}^{k}\Vert^{2},
\end{align*}}

\vspace{-2mm}
\noindent where parameters
{\footnotesize
\begin{align*}
&\hat{\theta}_{i}^{k+1}:=\frac{\omega_{i}^{k}}{2}-\frac{4}{\beta_{i}}(\mu(\psi_{0}\psi_{2}+\psi_{1}^{2}+(\overline{\mathcal{V}}_{i}+\overline{\mathcal{V}}_{i-1})\psi_{2})+\omega_{i}^{k})^{2}, i\in[N-1];\\
&\hat{\eta}_{i}:=\mu-\frac{4\mu^{2}\psi_{1}^{2}}{\beta_{i+1}}, i\in[N-2],~~ \hat{\eta}_{N-1}:=\frac{\mu}{2};\\
&\tilde{\theta}_{i}^{k}:=\frac{4(\omega_{i}^{k-1})^{2}}{\beta_{i}}, i\in[N-1];\\
&\tilde{\eta}_{i}:=\frac{4\mu^{2}\psi_{1}^{2}}{\beta_{i}}, i\in[N-1].
\end{align*}}

\noindent It can be verified that {\small$\frac{4(\underline{\omega}_i)^2}{\beta_i}+\frac{\underline{\omega}_i}{4}-\tilde{\theta}_{i}^{k}\geq\epsilon_i$}, {\small$\exists~\kappa_i>0$} such that {\small$\hat{\theta} _{i}^{k+1}-\tilde{\theta}_{i}^{k}-\frac{\omega_{i}^{k-1}}{4}>\kappa_i$}, {\small$\hat{\eta}_{i}-\tilde{\eta}_{i}-\frac{\mu}{4}>0, i\in[N-1]$}, and {\small$\frac{1+\beta_{N}}{2}-\frac{1}{\beta_{N}}>0$}.

\vspace{2mm}
\noindent Then, we have
\vspace{-2mm}
{\small
\begin{align*}
&\mathcal{L}_{\beta}^{3s}(\Psi_{3s}^{k+1})+\sum_{i=1}^{N-1}\theta_{i}\Vert \U_{i}^{k+1}-\U_{i}^{k}\Vert^{2}+\sum_{i=1}^{N-1}\eta_{i}\Vert \V_{i}^{k+1}-\V_{i}^{k}\Vert^{2}\\
\leq& \mathcal{L}_{\beta}^{3s}(\Psi_{3s}^{k})+\sum_{i=1}^{N-1}\theta_i\Vert \U_{i}^{k}-\U_{i}^{k-1}\Vert^{2}+\sum_{i=1}^{N-1}\eta_i\Vert \V_{i}^{k}-\V_{i}^{k-1}\Vert^{2}\\
&-\sum_{i=1}^{N-1}\kappa_i\Vert \U_{i}^{k+1}-\U_{i}^{k}\Vert^{2}-\sum_{i=1}^{N-1}(\hat{\eta}_{i}-\tilde{\eta}_{i}-\frac{\mu}{4})\Vert \V_{i}^{k+1}-\V_{i}^{k}\Vert^{2}\\
&-\sum_{i=1}^{N-1}\epsilon_i \Vert \U_{i}^{k}-\U_{i}^{k-1}\Vert^{2}-\sum_{i=1}^{N-1}\frac{\mu}{4}\Vert \V_{i}^{k}-\V_{i}^{k-1}\Vert^{2}\\
&-\frac{\lambda}{2}\sum_{i=1}^{N}\Vert \W_{i}^{k+1}-\W_{i}^{k}\Vert^{2}-(\frac{1+\beta_{N}}{2}-\frac{1}{\beta_{N}})\Vert \V_{N}^{k+1}-\V_{N}^{k}\Vert^{2}\\
\leq&\mathcal{L}_{\beta}^{3s}(\Psi_{3s}^{k})+\sum_{i=1}^{N-1}\theta_{i}\Vert \U_{i}^{k}-\U_{i}^{k-1}\Vert^{2}+\sum_{i=1}^{N-1}\eta_{i}\Vert \V_{i}^{k}-\V_{i}^{k-1}\Vert^{2}\\
&-\bar c\big(\sum_{i=1}^{N}\Vert \W_{i}^{k+1}-\W_{i}^{k}\Vert^{2}+\sum_{i=1}^{N-1}\Vert \U_{i}^{k+1}-\U_{i}^{k}\Vert^{2}\\
&+\sum_{i=1}^{N}\Vert \V_{i}^{k}-\V_{i}^{k-1}\Vert^{2}+\sum_{i=1}^{N-1}\Vert \U_{i}^{k-1}-\U_{i}^{k-2}\Vert^{2}\\
&+\sum_{i=1}^{N-1}\Vert \V_{i}^{k-1}-\V_{i}^{k-2}\Vert^{2}\big),
\end{align*}}
\vspace{-2mm}

\noindent where $\bar c=\min\{\kappa_i, \bar{\eta}_{i}-\tilde{\eta}_{i}-\frac{\mu}{4}, \epsilon_i, \frac{\mu}{4}, \frac{\lambda}{2}, \frac{\beta_{N}^2+\beta_{N}-2}{2\beta_{N}}\}>0$.
{\small\begin{align*}
&\sum_{i=1}^{N}\Vert\Lam_{i}^{k+1}-\Lam_{i}^{k}\Vert^{2}
\leq \tilde c\big(\sum_{i=1}^{N}(\Vert \W_{i}^{k+1}-\W_{i}^{k}\Vert^{2}+\Vert \V_{i}^{k+1}-\V_{i}^{k}\Vert^{2})\\
&+\sum_{i=1}^{N-1}(\Vert \U_{i}^{k+1}-\U_{i}^{k}\Vert^{2}+\Vert \V_{i}^{k}-\V_{i}^{k-1}\Vert^{2}+\Vert \U_{i}^{k}-\U_{i}^{k-1}\Vert^{2})\big)
\end{align*}}
\vspace{-1mm}

\noindent for certain $\tilde c>0$.
Therefore,
\vspace{-1mm}
{\small\begin{align*}
&\mathcal{L}_{\beta}^{3s}(\Psi_{3s}^{k+1})+\sum_{i=1}^{N-1}\theta_{i}\Vert \U_{i}^{k+1}-\U_{i}^{k}\Vert^{2}+\sum_{i=1}^{N-1}\eta_{i}\Vert \V_{i}^{k+1}-\V_{i}^{k}\Vert^{2},\\
\leq& \mathcal{L}_{\beta}^{3s}(\Psi_{3s}^{k})+\sum_{i=1}^{N-1}\theta_{i}\Vert \U_{i}^{k}-\U_{i}^{k-1}\Vert^{2}+\sum_{i=1}^{N-1}\eta_{i}\Vert \V_{i}^{k}-\V_{i}^{k-1}\Vert^{2}\\
&-\frac{\bar c}{2}\big(\sum_{i=1}^{N}\Vert \W_{i}^{k+1}-\W_{i}^{k}\Vert^{2}+\sum_{i=1}^{N-1}\Vert \U_{i}^{k+1}-\U_{i}^{k}\Vert^{2}\\
&+\sum_{i=1}^{N}\Vert \V_{i}^{k+1}-\V_{i}^{k}\Vert^{2}+\sum_{i=1}^{N-1}\Vert \U_{i}^{k}-\U_{i}^{k-1}\Vert^{2}\\
&+\sum_{i=1}^{N-1}\Vert \V_{i}^{k}-\V_{i}^{k-1}\Vert^{2}\big)-\frac{\bar c}{2\tilde c}\sum_{i=1}^{N}\Vert \Lam_{i}^{k+1}-\Lam_{i}^{k}\Vert^{2}\\
\leq& \mathcal{L}_{\beta}^{3s}(\Psi_{3s}^{k})+\sum_{i=1}^{N-1}\theta_{i}\Vert \U_{i}^{k}-\U_{i}^{k-1}\Vert^{2}+\sum_{i=1}^{N-1}\eta_{i}\Vert \V_{i}^{k}-\V_{i}^{k-1}\Vert^{2}\\
&-\frac{\bar c}{2\tilde c}\big(\sum_{i=1}^{N}\Vert \W_{i}^{k+1}-\W_{i}^{k}\Vert^{2}+\sum_{i=1}^{N-1}\Vert \U_{i}^{k+1}-\U_{i}^{k}\Vert^{2}\\
&+\sum_{i=1}^{N}\Vert \V_{i}^{k+1}-\V_{i}^{k}\Vert^{2}+\sum_{i=1}^{N}\Vert \Lam_{i}^{k+1}-\Lam_{i}^{k}\Vert^{2}\\
&+\sum_{i=1}^{N-1}\Vert \U_{i}^{k}-\U_{i}^{k-1}\Vert^{2}+\sum_{i=1}^{N-1}\Vert \V_{i}^{k}-\V_{i}^{k-1}\Vert^{2}\big),
\end{align*}}

\vspace{-1mm}
\noindent which means $\mathcal{L}_R^{3s}(\widetilde\Psi_{3s}^{k+1})\leq \mathcal{L}_R^{3s}(\widetilde\Psi_{3s}^{k})-\frac{\bar c}{2\tilde c}\Vert \widetilde\Psi_{3s}^{k+1}-\widetilde\Psi_{3s}^{k}\Vert^2$.
\end{IEEEproof}
\begin{lemma}\label{3s_prox_theorem1}
The serial and parallel three-splitting RADMMs satisfy \Cref{prop:2s_prox_theorem2} with respect to $\mathcal{L}_R^{3s}$ and $\widetilde\Psi_{3s}$.
\end{lemma}

\noindent In order to prove Lemma \ref{3s_prox_theorem1}, we first estimate the boundness lemma in \Cref{3s_prox_lemma9}.
\begin{lemma}\label{3s_prox_lemma9}
There exist positive constants $\overline{\mathcal{V}}_{N}$, $\{\overline{\mathcal{\lambda}}_{i}\}_{i=1}^{N}$, $\{\overline{\mathcal{W}}_{i}\}_{i=1}^{N}$, and $\{\overline{\mathcal{U}}_{i}\}_{i=1}^{N-1}$ such that $\Vert \V_{N}^{k}\Vert\leq\overline{\mathcal{V}}_{N}$, $\Vert\Lam_{i}^{k}\Vert\leq\overline{\mathcal{\lambda}}_{i}$, $\Vert \W_{i}^{k}\Vert\leq\overline{\mathcal{W}}_{i}$, $i\in[N]$, $\Vert \U_{i}^{k}\Vert\leq\overline{\mathcal{U}}_{i}$, $i\in[N-1]$.
\end{lemma}
\begin{IEEEproof}
Noting that for each $i\in [N-1]$,
{\small\begin{align*}
\Vert\Lam_{i}^{k}\Vert^{2}\leq &2\mu^{2} \psi_{1}^{2}(\sqrt{nd}\psi_{0}+\overline{\mathcal{V}}_{i-1}+\overline{\mathcal{V}}_{i})^{2}+2(\omega_{i}^{k-1})^{2}\Vert \U_{i}^{k}-\U_{i}^{k-1}\Vert^{2},
\end{align*}}

\noindent then $\mathcal{I}^k-\sum_{i=1}^{N-1}\frac{ \mu^{2}\psi_{1}^{2}}{\beta_{i}}(\sqrt{nd}\psi_{0}+\overline{\mathcal{V}}_{i-1}+\overline{\mathcal{V}}_{i})^{2}\le \mathcal{L}_R^{3s}(\widetilde\Psi_{3s}^{k})$, in which
\vspace{-3mm}
{\small
\begin{align*}
&\mathcal{I}^k:=\big(\frac{1}{2}-\frac{1}{2\beta_{N}}\big)\Vert \V_{N}^{k}-\Y\Vert^{2}+\frac{\lambda}{2}\sum_{i=1}^{N}\Vert \W_{i}^{k}\Vert^{2}\nonumber\\
&+\frac{\mu}{2}\sum_{i=1}^{N-1}\Vert \V_{i-1}^{k}+\sigma_{i}(\U_{i}^{k})-\V_{i}^{k}\Vert^{2}+\sum_{i=1}^{N-1}\eta_{i}\Vert \V_{i}^{k}-\V_{i}^{k-1}\Vert^{2}\nonumber \\
&+\sum_{i=1}^{N-1}\frac{\beta_{i}}{2}\big\Vert \W_{i}^{k}\V_{i-1}^{k}-\U_{i}^{k}+\frac{1}{\beta_{i}}\Lam_{i}^{k}\big\Vert^{2}\nonumber\\
&+\frac{\beta_{N}}{2}\big\Vert \W_{N}^{k}\V_{N-1}^{k}-\V_{N}^{k}+\frac{1}{\beta_{N}}\Lam_{N}^{k}\big\Vert^{2}\nonumber \\
&+\sum_{i=1}^{N-1}(\theta_{i}-\frac{\big(\omega_{i}^{k-1})^{2}}{\beta_{i}}\big)\Vert \U_{i}^{k}-\U_{i}^{k-1}\Vert^{2}.
\end{align*}}

\vspace{-1mm}

\noindent It can be verified that {\small$\theta_{i}-\frac{(\omega_{i}^{k-1})^{2}}{\beta_{i}}\geq\frac{3(\underline{\omega}_i)^2}{\beta_i}+\frac{3\underline{\omega}_i}{16}+\frac{\epsilon_i}{4}>0$, $i\in[N-1]$}.
By Lemma \ref{3s_prox_theorem1}, {\small$\{\mathcal{L}_R^{3s}(\widetilde\Psi_{3s}^{k})\}$} is upper bounded. If {\small$\Vert \V_{N}^{k}\Vert\to\infty$}, then {\small$\mathcal{L}_R^{3s}(\widetilde\Psi_{3s}^{k})\to\infty$}, a contradiction. Thus there exists {\small$\mathcal{V}_{N}>0$} such that {\small$\Vert \V_{N}^{k}\Vert\leq\overline{\mathcal{V}}_{N}$}. By {\small$\Lam_{N}^{k}=\V_{N}^{k}-\Y$}, there exists {\small$\overline{\mathcal{\lambda}}_{N}>0$} such that {\small$\Vert \Lam_{N}^{k}\Vert\leq\overline{\mathcal{\lambda}}_{N}$}. If {\small$\Vert \U_{i}^{k}-\U_{i}^{k-1}\Vert\to\infty$}, then {\small$\mathcal{L}_R^{3s}(\widetilde\Psi_{3s}^{k})\to\infty$}, a contradiction. Thus there exist {\small$\chi_{i}>0$} such that {\small$\Vert \U_{i}^{k}-\U_{i}^{k-1}\Vert\leq\chi_{i}$}, {\small$i\in[N-1]$}. According to {\small$\Lam_ {i}^{k}=\mu(\V_{i-1}^{k}+\sigma_i(\U_{i}^{k-1})-\V_{i}^{k-1})\odot\sigma_i^{'}(\U_{i}^{k-1})+\tau_{i}^{k-1}(\U_{i}^{k}-\U_{i}^{k-1})$}, there exist {\small$\overline{\mathcal{\lambda}}_{i}>0$} such that {\small$\Vert\Lam_{i}^{k}\Vert\leq\overline{\mathcal{\lambda}}_{i}$, $i\in[N-1]$}. If {\small$\Vert \W_{i}^{k}\Vert\to\infty$}, then {\small$\mathcal{L}_R^{3s}(\widetilde\Psi_{3s}^{k})\to\infty$}, a contradiction. Thus there exist {\small$\overline{\mathcal{W}}_{i}>0$} such that {\small$\Vert \W_{i}^{k}\Vert\leq\overline{\mathcal{W}}_{i}$}, {\small$i\in [N]$}.
By the following inequality:

\vspace{-1.5mm}
{\footnotesize\begin{align*}
&\sum_{i=1}^{N-1}\frac{\beta_{i}}{2}\big\Vert \W_{i}^{k}\V_{i-1}^{k}-\U_{i}^{k}+\frac{1}{\beta_{i}}\Lam_{i}^{k}\big\Vert^{2}\\
\geq&\sum_{i=1}^{N-1}\frac{\beta_{i}}{2}\big(\Vert \U_{i}^{k}\Vert-\overline{\mathcal{W}}_{i}\overline{\mathcal{V}}_{i-1}-\frac{1}{\beta_{i}}\overline{\mathcal{\lambda}}_{i}\big)^{2}
\end{align*}}

\noindent for sufficiently large $\Vert \U_{i}^{k}\Vert$. If $\Vert \U_{i}^{k}\Vert\to\infty$, then $\mathcal{L}(\widetilde\Psi_{3s}^{k})\to\infty$, a contradiction. Thus there exists $\overline{\mathcal{U}}_{i}>0$ such that $\Vert \U_{i}^{k}\Vert\leq\overline{\mathcal{U}}_{i}$, $i\in[N-1]$.
\end{IEEEproof}

\vspace{2mm}
\begin{IEEEproof}{(Proof of Lemma \ref{3s_prox_theorem1})}
With similar arguments as in the proof of Proposition \ref{prop:2s_prox_theorem2}, we have
{\small\begin{align*}
&\Vert\nabla \mathcal{L}_{\beta}^{3s}(\Psi_{3s}^{k+1})\Vert\\
\leq&\sum_{i=1}^{N}\Big\Vert\frac{\partial \mathcal{L}_{\beta}^{3s}}{\partial \W_{i}}(\Psi_{3s}^{k+1})\Big\Vert+\sum_{i=1}^{N}\Big\Vert\frac{\partial \mathcal{L}_{\beta}^{3s}}{\partial \V_{i}}(\Psi_{3s}^{k+1})\Big\Vert\\
&+\sum_{i=1}^{N-1}\Big\Vert\frac{\partial \mathcal{L}_{\beta}^{3s}}{\partial \U_{i}}(\Psi_{3s}^{k+1})\Big\Vert+\sum_{i=1}^{N}\Big\Vert\frac{\partial \mathcal{L}_{\beta}^{3s}}{\partial \Lam_{i}}(\Psi_{3s}^{k+1})\Big\Vert\\
\leq& \hat c\big(\sum_{i=1}^{N}(\Vert \W_{i}^{k+1}-\W_{i}^{k}\Vert+\Vert \V_{i}^{k+1}-\V_{i}^{k}\Vert+\Vert\Lam_{i}^{k+1}-\Lam_{i}^{k}\Vert)\\
&+\sum_{i=1}^{N-1}(\Vert \U_{i}^{k+1}-\U_{i}^{k}\Vert+\Vert \V_{i}^{k}-\V_{i}^{k-1}\Vert+\Vert \U_{i}^{k}-\U_{i}^{k-1}\Vert)\big)
\end{align*}}
\vspace{-2mm}

\noindent for certain $\hat c>0$. Thus
{\small\begin{align*}
&\Vert\nabla \mathcal{L}(\widetilde\Psi_{3s}^{k+1})\Vert\\
\leq&\Vert\nabla \mathcal{L}_{\beta}^{3s}(\Psi_{3s}^{k+1})\Vert+4\sum_{i=1}^{N-1}\big(\theta_{i}\Vert \U_{i}^{k+1}-\U_{i}^{k}\Vert+\eta_{i}\Vert \V_{i}^{k+1}-\V_{i}^{k}\Vert\big)\\
\leq &c_2\Vert\widetilde\Psi_{3s}^{k+1}-\widetilde\Psi_{3s}^{k}\Vert,
\end{align*}}

\noindent where $c_{2}=\sqrt{6N-3}(\hat c+4\max\{\{\theta_{i}\}_{i=1}^{N-1}, \{\eta_{i}\}_{i=1}^{N-1}\})$.
\end{IEEEproof}
\vspace{2mm}
\begin{IEEEproof}{(Proof of Lemma \ref{lem:1})}
By Lemmas \ref{lemma:3s_inter_result} and \ref{3s_prox_theorem1}, with similar arguments as in the proof of Theorems \ref{thm:thm_2s_prox_main_result1} and \ref{thm:3s_prox_theorem6}, we obtain the results.
\end{IEEEproof}
\vspace{-2mm}
\subsection{Proofs of Theorems \ref{thm:3} and \ref{thm:3s_prox_theorem6}}\label{subsection-theorem6-theorem7}
\begin{IEEEproof}{(Proof of Theorems \ref{thm:3} and \ref{thm:3s_prox_theorem6})}
We only need to establish the connections between $\{\mathcal{L}_\beta^{3s}, \Psi_{3s}\}$ and $\{\mathcal{L}_R^{3s}, \widetilde\Psi_{3s}\}$.
For the iteration point sequence,
\vspace{-1.5mm}
{\small\begin{align*}
&\widetilde\Psi_{3s}^{k}-\widetilde\Psi_{3s}^{*}\\
=&(\Psi_{3s}^{k}-\Psi_{3s}^{*}, \veccommand(\U_{i})^{k-1}-\veccommand(\U_{i})^{*}, \veccommand(\V_{i})^{k-1}-\veccommand(\V_{i})^{*}).
\end{align*}}

\vspace{-1mm}
\noindent Then we have $\Vert \Psi_{3s}^{k}-\Psi_{3s}^{*}\Vert^{2}\leq\Vert \widetilde\Psi_{3s}^{k}-\widetilde\Psi_{3s}^{*}\Vert^{2}$. Moreover,
\vspace{-1mm}
{\small\begin{align*}
\mathcal{L}_\beta^{3s}(\Psi_{3s}^*)=&\lim_{k\to\infty}\mathcal{L}_{\beta}^{3s}(\Psi_{3s}^k)+\lim_{k\to\infty}\sum_{i=1}^{N-1}\theta_{i}\Vert U_{i}^{k}-U_{i}^{k-1}\Vert^{2}\\
&+\lim_{k\to\infty}\sum_{i=1}^{N-1}\eta_{i}\Vert V_{i}^{k}-V_{i}^{k-1}\Vert^{2}=\mathcal{L}_{R}^{3s}(\widetilde\Psi_{3s}^*).
\end{align*}}

\vspace{-2mm}
\noindent Thus $\mathcal{L}_\beta^{3s}(\Psi_{3s}^k)-\mathcal{L}_\beta^{3s}(\Psi_{3s}^*)\le \mathcal{L}_{R}^{3s}(\widetilde\Psi_{3s}^k)-\mathcal{L}_{R}^{3s}(\widetilde\Psi_{3s}^*)$.

\end{IEEEproof}
\section{Acknowledgments}
The authors would like to thank Yau Mathematical Sciences Center, Tsinghua
University for providing the computing resources to support our research. We thank Dr. Xixi Jia for helpful suggestions.

\vspace{-3mm}
\bibliographystyle{ieeetr}
\bibliography{ref}

\vfill

\end{document}